\documentclass[final,5p,times,twocolumn]{elsarticle}

\usepackage{amsmath,amssymb,amsfonts}
\usepackage{algorithmic}
\usepackage{graphicx}
\usepackage{textcomp}
\usepackage[linesnumbered,ruled]{algorithm2e}
\usepackage{booktabs}
\usepackage{epsfig}
\usepackage{rotating}
\usepackage{subfigure}
\usepackage{caption,setspace}
\usepackage{xcolor}
\usepackage{stfloats}
\usepackage{placeins}

\let\vec\mathbf

\journal{Swarm and Evolutionary Computation}

\begin{document}

\begin{frontmatter}



\title{A Fast and Efficient Stochastic Opposition-Based Learning for Differential Evolution in Numerical Optimization}


\author[affil1,affil2]{Tae Jong Choi}
\author[affil1]{Julian Togelius}
\author[affil3]{Yun-Gyung Cheong}

\address[affil1]{Tandon School of Engineering, New York University, Brooklyn, NY 11201, USA}
\address[affil2]{Department of Electrical and Computer Engineering, Sungkyunkwan University, Suwon-si, Gyeonggi-do 16419, Republic of Korea}
\address[affil3]{College of Software, Sungkyunkwan University, Suwon-si, Gyeonggi-do 16419, Republic of Korea}

\begin{abstract}
A fast and efficient stochastic opposition-based learning (OBL) variant is proposed in this paper. OBL is a machine learning concept to accelerate the convergence of soft computing algorithms, which consists of simultaneously calculating an original solution and its opposite. Recently, a stochastic OBL variant called BetaCOBL was proposed, which is capable of controlling the degree of opposite solutions, preserving useful information held by original solutions, and preventing the waste of fitness evaluations. While it has shown outstanding performance compared to several state-of-the-art OBL variants, the high computational cost of BetaCOBL may hinder it from cost-sensitive optimization problems. Also, as it assumes that the decision variables of a given problem are independent, BetaCOBL may be ineffective for optimizing inseparable problems. In this paper, we propose an improved BetaCOBL that mitigates all the limitations. The proposed algorithm called iBetaCOBL reduces the computational cost from $O(NP^{2} \cdot D)$ to $O(NP \cdot D)$ ($NP$ and $D$ stand for population size and a dimension, respectively) using a linear time diversity measure. Also, the proposed algorithm preserves strongly dependent variables that are adjacent to each other using multiple exponential crossover. We used differential evolution (DE) variants to evaluate the performance of the proposed algorithm. The results of the performance evaluations on a set of 58 test functions show the excellent performance of iBetaCOBL compared to ten state-of-the-art OBL variants, including BetaCOBL.
\end{abstract}


\end{frontmatter}


\section{Introduction}
\label{sec:Introduction}

An evolutionary algorithm (EA) is a subset of evolutionary computation, which is a nature-inspired optimization technique. As an EA does not make any assumption, it can be applied to black-box optimization problems. An EA randomly initializes its individuals over the search space of a given problem and repeatedly updates them through evolutionary operators until a termination criterion is satisfied.

Differential evolution (DE) \cite{StandardDE-1, StandardDE-2} is a powerful EA for optimizing multidimensional real-valued functions. DE offers a straightforward implementation. Moreover, DE has shown outstanding performance in many competitions on numerical optimization \cite{DESurvey-2}. Furthermore, in contrast with covariance matrix adaptation evolutionary strategy (CMA-ES) \cite{CMA-ES} that is another powerful EA for optimizing multidimensional real-valued functions, DE can be applied to large-scale problems because of its low space complexity \cite{DESurvey-2}. DE has gathered much attention from researchers and practitioners for over two decades.

Since DE was introduced, numerous studies have been conducted to design new DE variants in an effort to improve performance \cite{DESurvey-1, DESurvey-2, DESurvey-3, DESurvey-4}. One of the successful branches within the studies is the combination of DE and opposition-based learning (OBL) \cite{OBL-1, OBL-2}. Inspired by the idea of opposite relationships among objects, OBL is a computational opposition concept designed to accelerate the convergence of soft computing algorithms, which consists of simultaneously calculating an original solution and its opposite. Despite its simplicity, OBL has successfully led to improvements in soft computing algorithms \cite{OBL-1, OBL-2, OBLSurvey-1, OBLSurvey-2, OBLSurvey-3}. The pioneering study on the combination of DE and OBL was conducted by Rahnamayan et al., resulting in opposition-based DE (ODE) \cite{ODE}. ODE runs OBL on population initialization and generation jumping, which calculates an original population and its opposite and merges them into one and selects the fittest individuals as population size.

Recently, a stochastic OBL variant called BetaCOBL was proposed \cite{BetaCODE}. BetaCOBL has three advantages over other OBL variants. First, it can control the degree of opposite solutions by using the convex and concave density functions adjusted by the beta distribution. Second, the partial dimensional change scheme of BetaCOBL is able to preserve useful information held by original solutions. Finally, the selection switching scheme of BetaCOBL is able to prevent the waste of fitness evaluations. BetaCOBL has shown outstanding performance compared to several state-of-the-art OBL variants \cite{BetaCODE}. However, the high computational cost of BetaCOBL may hinder it from cost-sensitive optimization problems. Also, as it assumes that the decision variables of a given problem are independent, BetaCOBL may be ineffective for optimizing inseparable problems.

In this paper, we propose an improved BetaCOBL that mitigates all the limitations. Instead of using a power mean-based diversity measure \cite{powerMeanBased-1, powerMeanBased-2} in the selection switching scheme we employed a linear time diversity measure \cite{allPossiblePoints-1, allPossiblePoints-2, allPossiblePoints-3, allPossiblePoints-4, classical-1, classical-2, classical-3, allPossiblePoints-linearTime-1, allPossiblePoints-linearTime-2} to reduce the computational cost. We found that, regarding the diversity measure, replacing the power mean by the linear time maintains the performance of BetaCOBL with considerably less time complexity. Also, instead of using binomial crossover in the partial dimensional change scheme, we employed multiple exponential crossover \cite{MER} to preserve strongly dependent variables that are adjacent to each other. We carried out experiments on the IEEE Congress of Evolutionary Computation (CEC) 2013 and 2017 test suites \cite{CEC2013, CEC2017}. We used three DE variants, DE/rand/1/bin, EDEV \cite{EDEV}, and LSHADE-RSP \cite{LSHADE-RSP}, to evaluate the performance of the proposed algorithm. The results of the performance evaluations on a set of 58 test functions show the excellent performance of iBetaCOBL compared to ten state-of-the-art OBL variants, including BetaCOBL. Notably, compared to its predecessor BetaCOBL, iBetaCOBL is competitive with considerably less time complexity.

The main contributions of this paper are as follows.

\begin{enumerate}
\item A new stochastic OBL variant called iBetaCOBL is proposed, which is competitive with ten state-of-the-art OBL variants. \label{item:1}
\item iBetaCOBL significantly outperforms its predecessor BetaCOBL with considerably less time complexity. \label{item:2}
\item iBetaCOBL can be readily embedded into any DE variant as a module. \label{item:3}
\end{enumerate}

The remainder of this paper is organized as follows: We introduce the fundamentals of DE and OBL in Section \ref{sec:Background}. In Section \ref{sec:RelatedWork}, we present several state-of-the-art OBL variants, especially for their development. In Section \ref{sec:ProposedAlgorithm}, the details of the proposed algorithm will be discussed after first reviewing BetaCOBL, which is the basis of the proposed algorithm. We introduce the experimental setup in Section \ref{sec:ExperimentalSetup}. We present the results of the performance evaluations in Sections \ref{sec:ResultsAndComparisons} and \ref{sec:PerformanceEnhancementOfDEVariants}. Finally, we conclude this paper in Section \ref{sec:Conclusion}.


\section{Background}
\label{sec:Background}

\subsection{Differential Evolution}

DE \cite{StandardDE-1, StandardDE-2} is a powerful EA for optimizing multidimensional real-valued functions; it involves having a population of $NP$ individuals. Each individual is a $D$-dimensional vector denoted by $\vec{x}_{i,g} = ( x_{i,g}^{1}, x_{i,g}^{2}, \cdots, x_{i,g}^{D} )$ where $g$ stands for a generation. At the beginning of an optimization process, DE randomly distributes the population over the search space of a given problem. The individuals explore the search space through evolutionary operators. If an individual finds a new location with a better fitness value, the individual moves to the location; otherwise, it stays. DE consists of four operators: 1) initialization, 2) mutation, 3) crossover, and 4) selection. We briefly introduce the operators in the following subsections.

\subsubsection{Initialization}

The role of the initialization operator is to randomly distribute the population over the search space of a given problem. Let the minimum and maximum bounds be $\vec{x}_{min} = ( x_{min}^{1}, x_{min}^{2}, \cdots, x_{min}^{D} )$ and $\vec{x}_{max} = ( x_{max}^{1}, x_{max}^{2}, \cdots, x_{max}^{D} )$, respectively. Each individual is initialized according to

\begin{equation}
x_{i,0}^{j} = x_{min}^{j} + rand_{i,j} \cdot ( x_{max}^{j} - x_{min}^{j} )
\end{equation}

\noindent
where $rand_{i,j}$ stands for a uniformly distributed random number within the $[0,1]$ range.

\subsubsection{Mutation}

The role of the mutation operator is to generate a set of mutant vectors. The mutant vector $\vec{v}_{i,g}$ is generated by using a linear combination of the three donor vectors, $\vec{x}_{r_{1},g}$, $\vec{x}_{r_{2},g}$, and $\vec{x}_{r_{3},g}$. The donor vectors are randomly selected from the population, mutually exclusive, and distinct from the target vector $\vec{x}_{i,g}$. Each mutant vector is formed according to

\begin{equation}
\vec{v}_{i,g} = \vec{x}_{r_{1},g} + F \cdot ( \vec{x}_{r_{2},g} - \vec{x}_{r_{3},g} )
\end{equation}

\noindent
where $F$ stands for a scaling factor that controls the scale of the difference $( \vec{x}_{r_{2},g} - \vec{x}_{r_{3},g} )$.

\subsubsection{Crossover}

The role of the crossover operator is to generate a set of trial vectors. The trial vector $\vec{u}_{i,g}$ is generated by recombining the mutant and target vectors, $\vec{v}_{i,g}$ and $\vec{x}_{i,g}$. Let the random index be $j_{rand} \in \{ 1, 2, \cdots, D \}$. Each trial vector is formed according to

\begin{equation}
u_{i,g}^{j} = \left\{ \begin{array}{ll}
v_{i,g}^{j} & \textrm{if $rand_{i,j} \leq CR$ or $j == j_{rand}$} \\
x_{i,g}^{j} & \textrm{otherwise}
\end{array} \right.
\end{equation}

\noindent
where $CR$ stands for a crossover rate that controls the rate between the mutant and target vectors.

\subsubsection{Selection}

The selection operator compares the fitness value of the trial and target vectors and picks the better one for the next generation. If the trial vector $\vec{u}_{i,g}$ has a better fitness value than the target vector $\vec{x}_{i,g}$, the trial vector is selected, and the target vector is discarded; otherwise, vice versa. Each individual for the next generation is formed according to

\begin{equation}
\vec{x}_{i,g+1} = \left\{ \begin{array}{ll}
\vec{u}_{i,g} & \textrm{if $f(\vec{u}_{i,g}) \leq f(\vec{x}_{i,g})$} \\
\vec{x}_{i,g} & \textrm{otherwise.}
\end{array} \right.
\end{equation}

\noindent
where $f(\vec{x})$ stands for an objective function to be minimized.

\subsubsection{Advanced Differential Evolution Variants}

Since DE was introduced, numerous studies have been conducted to design new DE variants in an effort to improve performance, such as adaptive trial vector generation strategies \cite{adaptiveStrategy-1, adaptiveStrategy-2, adaptiveStrategy-3, adaptiveStrategy-4, adaptiveStrategy-5, adaptiveStrategy-6}, adaptive parameter controls \cite{adaptiveControlParameter-1, adaptiveControlParameter-2, adaptiveControlParameter-3, adaptiveControlParameter-4, adaptiveControlParameter-5, adaptiveControlParameter-6, adaptiveControlParameter-7, adaptiveControlParameter-8}, ensemble techniques \cite{EPSDE, MPEDE, EDEV}, and incorporating external techniques, such as $\alpha$-stable distribution based trial vector generation strategies \cite{Cauchy-1, Cauchy-2, Cauchy-3, Cauchy-4, Cauchy-5}, neighborhood-based trial vector generation strategies \cite{neighborhoodBased}, and OBLs \cite{ODEVariants-1, ODEVariants-2, ODEVariants-3, ODEVariants-4, ODEVariants-5, ODEVariants-6, ODEVariants-7, ODEVariants-8}. For more detailed explanations of state-of-the-art DE variants, please refer to the following surveys \cite{DESurvey-1, DESurvey-2, DESurvey-3, DESurvey-4}.

\subsection{Opposition-Based Learning}

Inspired by the idea of opposite relationships among objects, Tizhoosh \cite{OBL-1} proposed a computational opposition concept called OBL, which consists of simultaneously calculating an original solution and its opposite. Despite its simplicity, OBL has proven to be effective in improving soft computing algorithms, such as artificial neural networks, EAs, fuzzy logic, and reinforcement learning \cite{OBL-1, OBL-2, OBLSurvey-1, OBLSurvey-2, OBLSurvey-3}. Also, it was mathematically proved that opposite values are more likely to be located near the optimal solution of a given problem than random values \cite{}.

An opposite solution in an one-dimensional space can be defined as follows.

\begin{quote}
\textbf{Definition 1} \cite{OBL-1}: Let the original solution be $x \in [ x_{min}, x_{max} ]$. The opposite solution for $x$ denoted by $\breve{x}$ is obtained as follows:

\begin{equation}
\breve{x} = x_{min} + x_{max} - x
\end{equation}
\end{quote}

\noindent
Similarly, an opposite solution in a $D$-dimensional space can be defined as follows.

\begin{quote}
\textbf{Definition 2} \cite{OBL-1}: Let the original solution be $\vec{x} = ( x^{1}, x^{2}, \cdots, x^{D} )$, $x^{j} \in [ x_{min}^{j}, x_{max}^{j} ]$. The opposite solution for $\vec{x}$ denoted by $\breve{\vec{x}} = ( \breve{x}^{1}, \breve{x}^{2}, \cdots, \breve{x}^{D} )$ is obtained as follows:

\begin{equation}
\breve{x}^{j} = x_{min}^{j} + x_{max}^{j} - x^{j}
\end{equation}
\end{quote}

\noindent
The opposite solution is the type-I opposition. It is the type-II opposition if an opposite solution in a $D$-dimensional space is calculated in the objective space of a given problem, which can be defined as follows.

\begin{quote}
\textbf{Definition 3} \cite{OBL-2}: Let the objective function be $f(\vec{x})$, $y_{min} \leq f(\vec{x}) \leq y_{max}$. Also, let the original solution be $\vec{x} = ( x^{1}, x^{2}, \cdots, x^{D} )$, $x^{j} \in [ x_{min}^{j}, x_{max}^{j} ]$. The opposite solution for $\vec{x}$ denoted by $\breve{\vec{x}} = ( \breve{x}^{1}, \breve{x}^{2}, \cdots, \breve{x}^{D} )$ is obtained as follows:

\begin{equation}
\breve{\vec{x}} = {\{ \vec{c} \mid \breve{y} = y_{min} + y_{max} - f(\vec{x}) \}}
\end{equation}
\end{quote}

\noindent
It should be noted that the type-II opposition requires the prior knowledge of the objective space of a given problem. Therefore, it is difficult to apply the type-II opposition to black-box optimization problems. Finally, OBL can be defined as follows:

\begin{quote}
\textbf{Definition 4} \cite{OBL-1}: Let the original and opposite solutions be $\vec{x} = ( x^{1}, x^{2}, \cdots, x^{D} )$ and $\breve{\vec{x}} = ( \breve{x}^{1}, \breve{x}^{2}, \cdots, \breve{x}^{D} )$, respectively. OBL selects the opposite solution if $f(\breve{\vec{x}}) \leq f(\vec{x})$; otherwise, vice versa.
\end{quote}


\section{Related Work}
\label{sec:RelatedWork}

Since the implementation of OBL, numerous studies have been carried out to design new variants of OBL in an effort to improve performance. In this section, we describe several state-of-the-art OBL variants.

As researchers and practitioners have actively embedded OBL variants into DE \cite{OBLSurvey-2}, numerous OBL variants have been proposed in the form of ODE variants. The pioneering study on the combination of DE and OBL was conducted by Rahnamayan et al., resulting in opposition-based DE (ODE) \cite{ODE}. To automatically tune the jumping rate, Rahnamayan et al. proposed an ODE variant called ODE with time-varying jumping rates (ODETVJRs) and found that a linearly decreasing jumping rate is more effective than a linearly increasing \cite{ODETVJR}. To prevent the waste of fitness evaluations, Esmailzadeh and Rahnamayan proposed an ODE variant called ODE with protective generation jumping (ODEPGJ), which stops OBL if the success rate of opposite solutions decreases in a row for a predefined threshold \cite{ODEPGJ}. In \cite{QODE}, quasi OBL (QOBL) was proposed, which searches for quasi opposite solutions between the center point and a given original solution. In \cite{QRODE} quasi reflection OBL (QROBL) was proposed, which searches for quasi reflection opposite solutions between the center point and the opposite solution of a given original solution. In \cite {COODE}, current-optimum-based ODE (COODE) was proposed, which uses the location of the current-optimum as a reference point to calculate opposite solutions. In \cite{GODE}, generalized ODE (GODE) was proposed, which uses a dynamically scaled search space and a uniformly distributed random number as a reference point. Zhou et al. proposed an extension of GODE called elite ODE (EODE), which calculates opposite solutions with the elite individuals. \cite{EODE}. Liu et al. proposed another extension of GODE called adaptive GODE (AGODE), which automatically tunes the jumping rate based on the success rate of opposite solutions \cite{AGODE}.

For more detailed explanations of state-of-the-art OBL variants, please refer to the following surveys \cite{OBLSurvey-1, OBLSurvey-2, OBLSurvey-3}.


\section{Proposed Algorithm}
\label{sec:ProposedAlgorithm}

The proposed algorithm, namely iBetaCOBL, is introduced in this section. The details of the modified schemes will be discussed after first reviewing BetaCOBL \cite{BetaCODE}, which is the basis of the proposed algorithm.

\subsection{Review of BetaCOBL}
\label{sec:ProposedAlgorithm_ReviewOfBetaCOBL}

\begin{figure}[h]
\includegraphics[width=\linewidth]{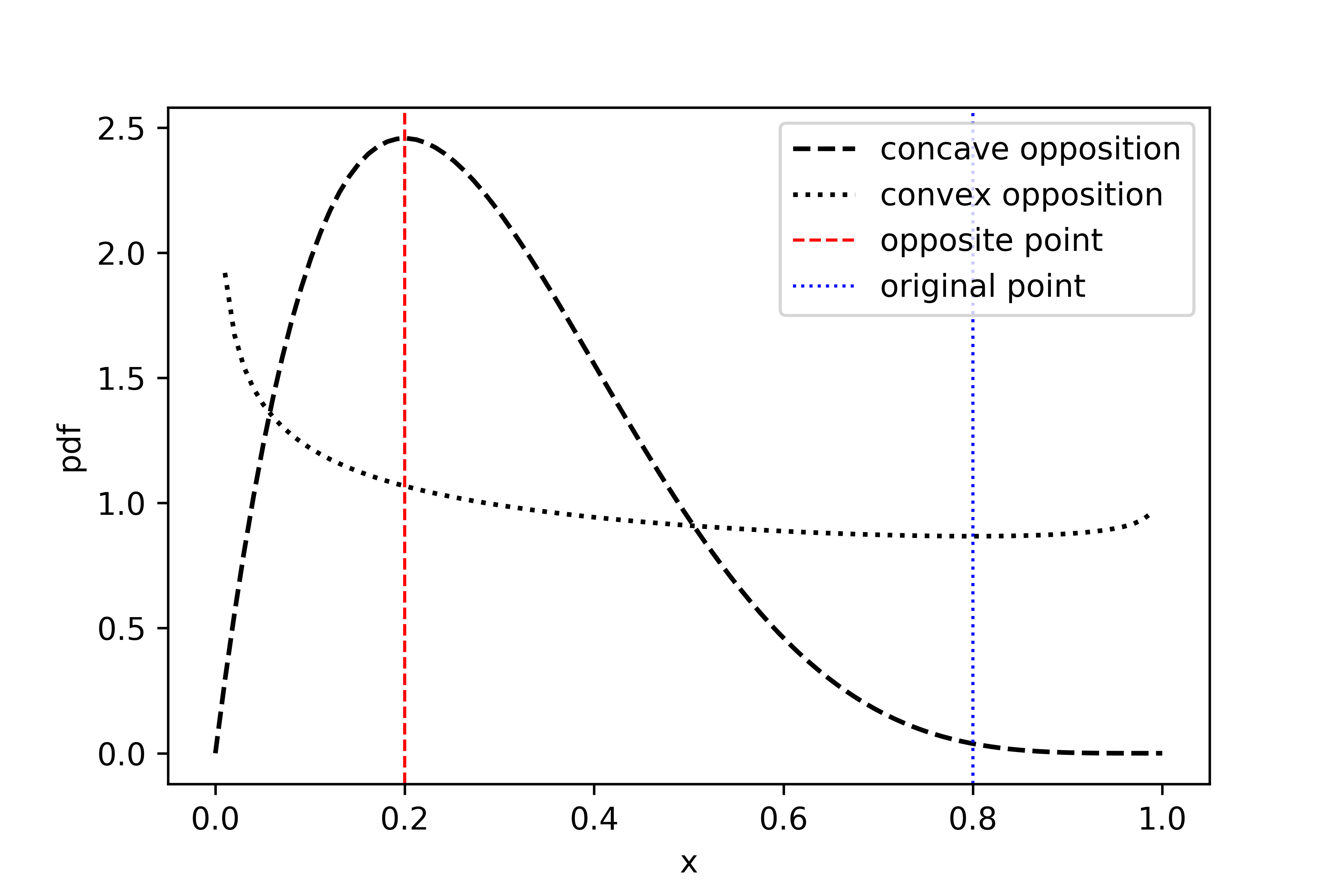}
\caption{Example of concave and convex opposite points}
\centering
\label{fig:exampleOfConcaveAndConvexOppositePoints}
\end{figure}

The following drawbacks affect numerous OBL variants:
1) As OBL variants compute opposite solutions or based on the uniform distribution, there is an inherent limitation in the deterministically search for decent opposite solutions. In other words, there is an opportunity for improvement when computing opposite solutions by using useful probability distributions, such as Cauchy, Gaussian, and $\alpha$-stable ones.
2) When OBL variants compute opposite solutions, the useful elements held with the original solutions can be discarded as all of the elements of the original solutions are transformed into opposites.
3) As OBL variants follow a greedy strategy, fitness evaluations can be wasted if suitable opposite solutions can no longer be discovered at the end of the optimization process.

To overcome these limitations, BetaCOBL uses the following techniques: 1) Beta distribution: BetaCOBL calculates concave or convex opposite solutions by using the beta distribution, which can create various shapes for the continuous probability density functions (PDFs) within the range $[0,1]$. Here, a concave opposite solution represents a solution generated based on a PDF where the opposite point for a given original solution is selected with the highest probability. Conversely, a convex opposite solution is generated based on a PDF where the point for a given original solution is selected with the lowest probability. As a result, with the concave and convex OBLs, BetaCOBL can find appropriate opposite solutions faster than other OBL variants. Fig. \ref{fig:exampleOfConcaveAndConvexOppositePoints} shows an example of concave and convex opposite points.

2) Partial dimensional change scheme: BetaCOBL uses the binomial crossover in DE to calculate a partial opposite solution, formed by the recombination of an original solution and its complete opposite solution. Therefore, BetaCOBL can obtain more diverse opposite solutions than other OBL variants as it can have one of the $2^{D}$ possible opposite solutions with a given pair of original and complete opposite solutions. In addition, as it uses the binomial crossover, BetaCOBL can preserve the useful elements held by original solutions.

3) Selection switching scheme: In general, OBL helps discover promising regions at the beginning of an optimization process, but it becomes less effective as the optimization process progresses; as a result, fitness evaluations are potentially wasted. To mitigate this issue, BetaCOBL estimates the population diversity before the concave and convex OBLs. If the population diversity is higher than a predefined threshold $DT$, BetaCOBL uses a $(\mu + \lambda)$ selection with all the original solutions of the population; otherwise, it uses a $(\mu, \lambda)$ selection with the worst half original solutions of the population. Consequently, BetaCOBL can prevent the waste of fitness evaluations by applying one of the two selection operators depending on the convergence progress.

A concave opposite solution is calculated using the beta distribution with both $\alpha$ and $\beta$ greater than one, as follows:

\begin{equation}
\breve{x}_{i,g}^{j} = (x_{max}^{j} - x_{min}^{j}) \cdot Beta(\alpha, \beta) + x_{min}^{j}
\label{eqn:oppositeSolution}
\end{equation}

\begin{equation}
\alpha = \left\{ \begin{array}{ll}
spread \cdot peak & \textrm{if $mode < 0.5$} \\
spread & \textrm{otherwise}
\end{array} \right.
\label{eqn:alpha}
\end{equation}

\begin{equation}
\beta = \left\{ \begin{array}{ll}
spread & \textrm{if $mode < 0.5$} \\
spread \cdot peak & \textrm{otherwise}
\end{array} \right.
\label{eqn:beta}
\end{equation}

\begin{equation}
spread = \big( \frac{1}{\sqrt{normDiv}} \big)^{1 + N(0, 0.5)}
\label{eqn:concaveSpread}
\end{equation}

\begin{equation}
peak = \left\{ \begin{array}{ll}
\frac{(spread - 2) \cdot mode + 1}{spread \cdot (1 - mode)} & \textrm{if $mode < 0.5$} \\
\frac{2 - spread}{spread} + \frac{spread - 1}{spread \cdot mode} & \textrm{otherwise}
\end{array} \right.
\label{eqn:peak}
\end{equation}

\begin{equation}
mode = \frac{(x_{min}^{j} + x_{max}^{j} - x_{i,g}^{j}) - x_{min}^{j}}{x_{max}^{j} - x_{min}^{j}}
\label{eqn:concaveMode}
\end{equation}

\noindent
where $Beta(\alpha, \beta)$ and $N(0, 0.5)$ denote the beta distribution with parameters $\alpha$ and $\beta$, and the Gaussian distribution with the mean $0$ and variance $0.5$, respectively. In addition, the normalized diversity denoted by $normDiv$ is calculated as follows:

\begin{equation}
normDiv = \frac{1}{NP} \sum_{i=1}^{NP} CD(\vec{x}_{i,g}, \mathbf{P}_{g})
\label{eqn:normDiv}
\end{equation}

\begin{equation}
CD(\vec{x}_{i,g}, \mathbf{P}_{g}) = \min_{\vec{c} \in \mathbf{P}_{g}, \vec{c} \ne \vec{x}_{i,g}} d(\vec{c}, \vec{x}_{i,g})
\label{eqn:CD}
\end{equation}

\begin{equation}
d(\vec{c}, \vec{x}_{i,g}) = \sqrt{\frac{1}{D} \sum_{j=1}^{D} \big( \frac{x_{i,g}^{j} - c^{j}}{x_{max}^{j} - x_{min}^{j}} \big)^{2}}
\label{eqn:d}
\end{equation}

The same formulas calculate a convex opposite solution except for the mode and spread, calculated as follows:

\begin{equation}
mode = \frac{x_{i,g}^{j} - x_{min}^{j}}{x_{max}^{j} - x_{min}^{j}}
\label{eqn:convexMode}
\end{equation}

\begin{equation}
spread = 0.1 \cdot \sqrt{normDiv} + 0.9
\label{eqn:convexSpread}
\end{equation}


\subsection{Modified Selection Switching Scheme}
\label{sec:ProposedAlgorithm_ModifiedSelectionSwitchingScheme}

\subsubsection{Problem of Selection Switching Scheme}
\label{sec:Problem1}

BetaCOBL uses the selection switching scheme to prevent the waste of fitness evaluations, which applies one of the two selection operators depending on the population diversity. To estimate the population diversity, BetaCOBL calculates the average of the minimum distance between all possible pairs, which is a power mean-based diversity measure. A generalized definition of the power mean-based diversity measure is presented in \cite{powerMeanBased-1, powerMeanBased-2}, where it is defined as the mapping $D_{h} : {\rm I\!R}^{NP \times D} \rightarrow {\rm I\!R}$

\begin{equation}
D_{h}(\mathbf{P}_{g}, a, b) = \sqrt[b]{\frac{1}{NP} \sum_{i=1}^{NP} d_{i}^{a}}
\label{eqn:D_h}
\end{equation}

\begin{equation}
d_{i}^{a} = \frac{1}{NP-1} \sum_{j=1}^{NP} \| \vec{x}_{i,g} - \vec{x}_{j,g} \|^{a}
\end{equation}

\noindent
where $a, b \neq 0$. The two parameters $a$ and $b$ determine the behavior of the diversity measure. If $a = 1$, the arithmetic mean distance of all possible pairs is computed. If $a = 0$, the geometric mean distance of all possible pairs is computed. In addition, if $a = -\infty$, the diversity measure evaluates the minimum distance of all possible pairs. Finally, the lower the value of $a$ and $b$, the larger the penalty to the collocation of individuals. The power mean-based diversity measure with $a = -\infty$ and $b = 1$ that BetaCOBL uses was experimentally proven not to be $(\rho, \epsilon)$-ectropy where both $\rho$ and $\epsilon$ can simultaneously take values close to zero \cite{ectropy}, which means it can discourage the collocation of individuals.

However, the power mean-based diversity measure with $a = -\infty$ and $b = 1$ incurs a $O(NP^{2} \cdot D)$ computational cost; as a result it is difficult to use BetaCOBL for optimizing more complex problems with a large population size.

\subsubsection{Applying Linear Time Diversity Measure}
\label{sec:ProposedAlgorithm_ModifiedSelectionSwitchingScheme_2}

To reduce the computational cost, we replaced the power mean-based diversity measure with a linear time diversity measure in the selection switching scheme. Of the two well-known measures, we employed one that computes the arithmetic mean of the Euclidean distances of all possible pairs \cite{allPossiblePoints-1, allPossiblePoints-2, allPossiblePoints-3, allPossiblePoints-4, classical-1, classical-2, classical-3, allPossiblePoints-linearTime-1, allPossiblePoints-linearTime-2}, where it can be defined as the mapping $D_{d} : {\rm I\!R}^{NP \times D} \rightarrow {\rm I\!R}$

\begin{equation}
D_{d}(\mathbf{P}_{g}) = \frac{1}{2} \sum_{i=1}^{n} \sum_{j=1}^{n} \| \vec{x}_{i,g} - \vec{x}_{j,g} \|
\label{eqn:D_d}
\end{equation}

\noindent
A naive implementation for equation (\ref{eqn:D_d}) incurs a $O(NP^{2} \cdot D)$ computational cost. Wineberg and Oppacher \cite{allPossiblePoints-linearTime-1, allPossiblePoints-linearTime-2} reformulated the equation for a linear time diversity measure as follows:

\begin{equation}
D_{d}'(\mathbf{P}_{g}) = \frac{1}{D} \sqrt{\sum_{k=1}^{D} \overline{(x_{g}^{k})^{2}} - (\overline{x_{g}^{k}})^{2}}
\label{eqn:D_d'}
\end{equation}

\noindent
where $\overline{(x_{g}^{k})^2} = \frac{1}{NP} \sum_{i=1}^{NP} (x_{i,g}^{k})^{2}$ and $\overline{x_{g}^{k}} = \frac{1}{NP} \sum_{i=1}^{NP} x_{i,g}^{k}$. The computational cost of the reformulated diversity measure is $O(NP \cdot D)$. Note that the proposed algorithm uses the normalized version of the diversity measure, obtained by dividing $\overline{(x_{g}^{k})^{2}} - (\overline{x_{g}^{k}})^{2}$ by $(x_{max}^{k} - x_{min}^{k})$ in the equation (\ref{eqn:D_d'}).

\subsubsection{Rationale of Employing Linear Time Diversity Measure}
\label{sec:ProposedAlgorithm_ModifiedSelectionSwitchingScheme_3}

\begin{figure}
\centering
\subfigure[All possible pair]{\includegraphics[scale=0.225]{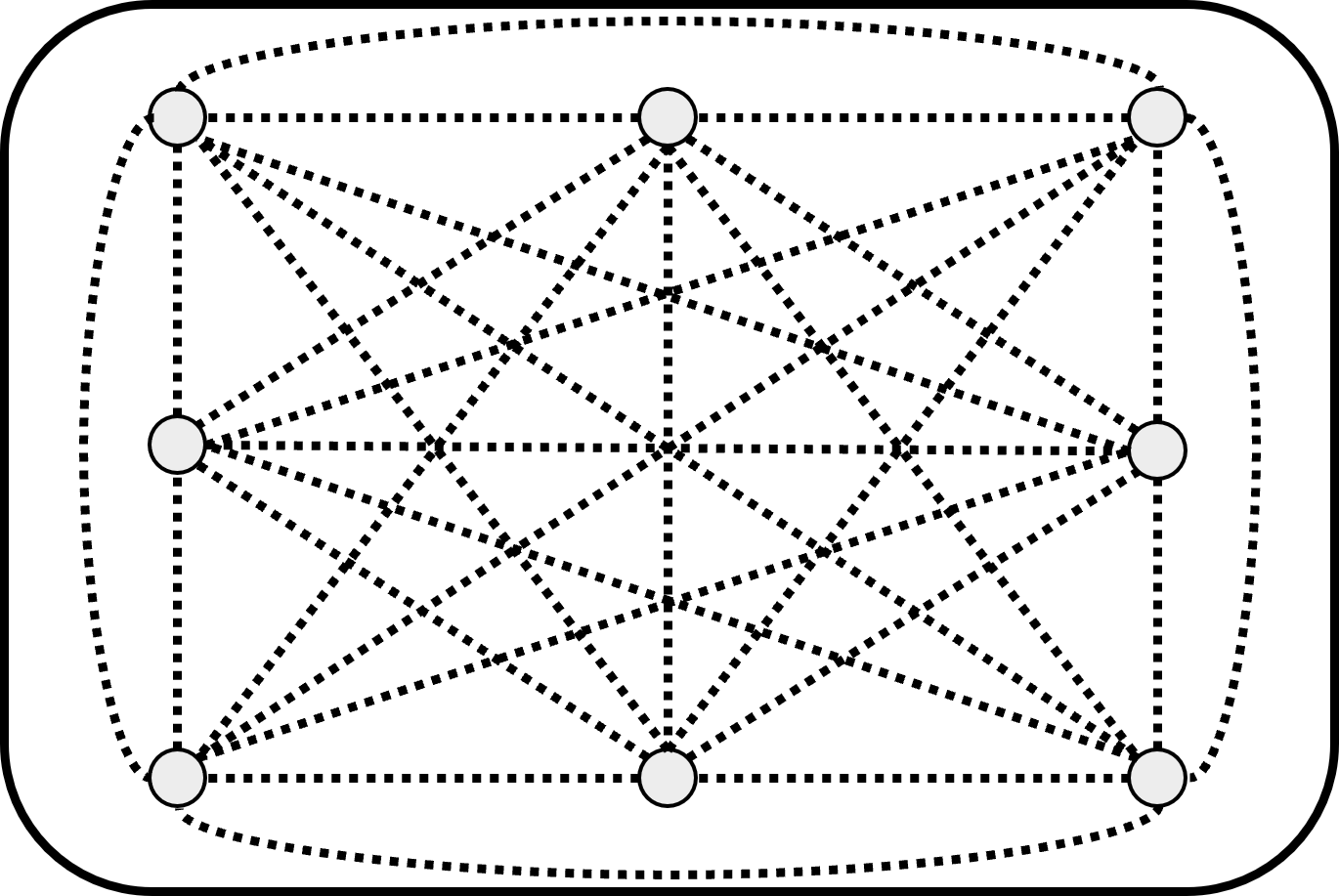}}\quad
\subfigure[Every point to center point]{\includegraphics[scale=0.225]{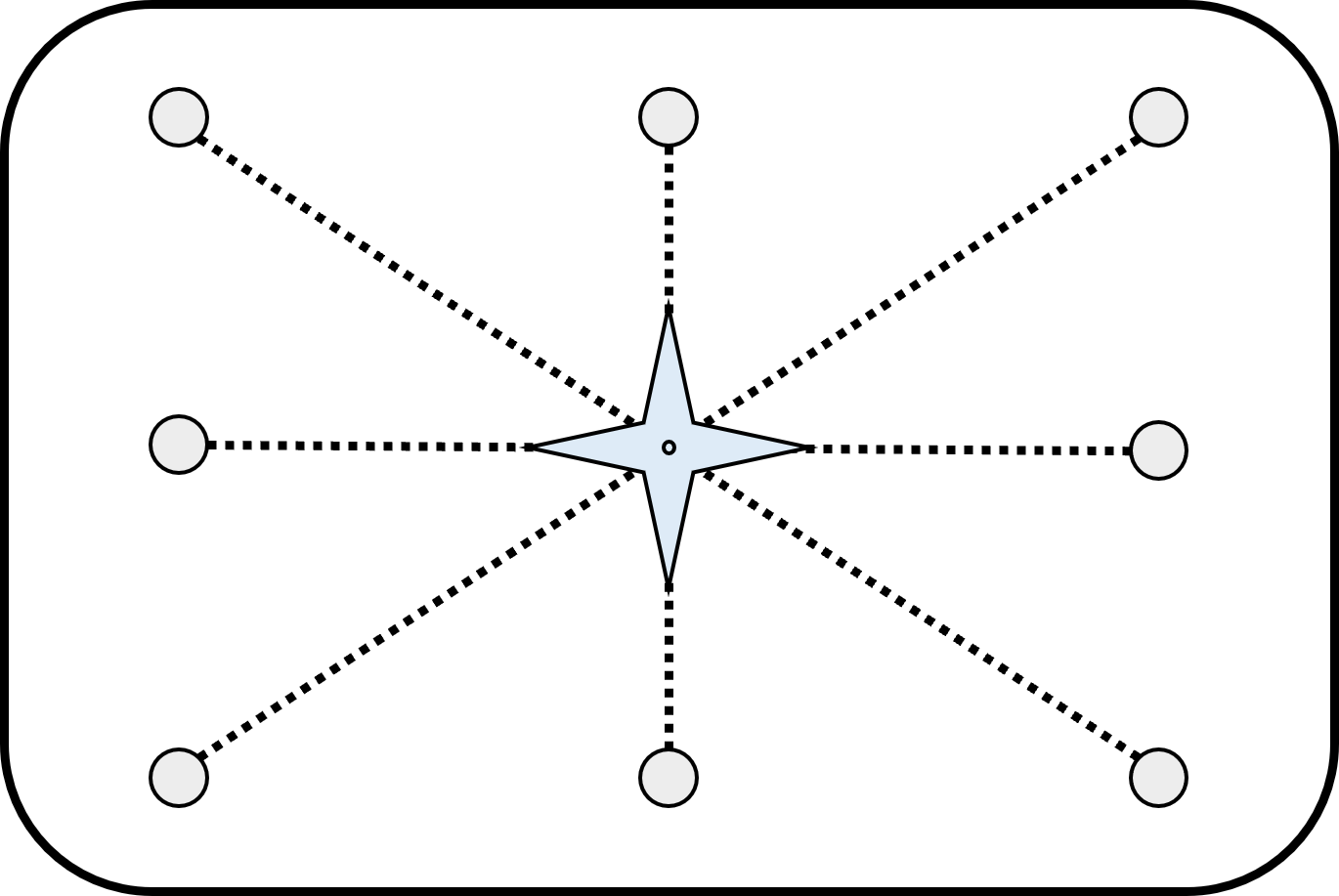}}
\caption{Two linear time diversity measures}
\label{fig:twoLinearTimeDiversityMeasures}
\end{figure}

As mentioned in Section \ref{sec:Problem1}, BetaCOBL uses the power mean-based diversity measure to check the convergence progress, which leads to a high computational cost. Therefore, we must replace it with a fast diversity measure to apply BetaCOBL to more complex problems with a large population size.

There are two linear time diversity measures in the multidimensional continuous space. The first measure was discussed in Section \ref{sec:ProposedAlgorithm_ModifiedSelectionSwitchingScheme_2} and the other computes the arithmetic mean of the Euclidean distances of every point to the center \cite{centerBased-1, centerBased-2, centerBased-3, centerBased-4, classical-1, classical-2, classical-3}, where it can be defined as the mapping $D_{v} : {\rm I\!R}^{NP \times D} \rightarrow {\rm I\!R}$

\begin{equation}
D_{v}(\mathbf{P}_{g}) = \sum_{i=1}^{n} \| \vec{x}_{i,g} - \overline{\vec{x}}_{g} \|
\label{eqn:D_v}
\end{equation}

\noindent
where $\overline{\vec{x}}_{g} = (M^{1}, M^{2}, \cdots, M^{D})$ and the centroid of the population with $M^{k} = \frac{1}{NP} \sum_{i=1}^{NP} x_{i,g}^{k}$, $k = 1, 2, \cdots, D$. Fig. \ref{fig:twoLinearTimeDiversityMeasures} shows the two linear time diversity measures.

We chose the first measure as it was theoretically proven to discourage the collocation of individuals bigger than the second measure \cite{ectropy}. Let the population size for each measure be $NP = 2^{m} + p$. The ectropic property of the first measure $D_{d}$ is $(\frac{2^{m}}{2^{m}+p},0)$, while that of the second measure $D_{v}$ is $(\frac{1}{2^{m}+p},0)$. Therefore, in a situation where many individuals are in overlapping positions, the second measure is more likely to return a higher value than the first one. In other words, BetaCOBL with the second measure is likely to continue to use the $(\lambda + \mu)$ selection instead of $(\lambda, \mu)$ at the end of the optimization process, which may not prevent the waste of fitness evaluations.

Consequently, the proposed algorithm can estimate the population diversity faster than BetaCOBL with the replacement. In addition, we analyze the relative performance of the original BetaCODE and BetaCODE with the linear time diversity measure and found that there were no significant differences, as reported in Section \ref{sec:ResultsAndComparisons}.


\subsection{Modified Partial Dimensional Change Scheme}
\label{sec:ProposedAlgorithm_ModifiedPartialDimensionalChangeScheme}

\subsubsection{Problem of Partial Dimensional Change Scheme}
\label{sec:Problem2}

BetaCOBL uses the binomial crossover in the partial dimensional change scheme to calculate partial opposite solutions. The binomial crossover is the most frequently used crossover operator in DE literature, and has the following properties \cite{crossoverStudy, MER}. First, the relationship between the mutation probability \cite{mutationProbability} and the control parameter $CR$ is linear. Second, the binomial crossover can generate all the $2^{D}$ possible trial vectors with a given pair of target and mutant vectors. However, it assumes that decision variables are not inter-related; thus, it tends to split up strongly dependent decision variables.

The exponential crossover is the traditional alternative crossover operator; it can preserve adjacent decision variables because of its sequential construct. Although this property helps search for decent solutions on inseparable problems, it has the following critical limitations \cite{crossoverStudy, MER}. First, the control parameter $CR$ is difficult to tune as the relationship between the mutation probability and $CR$ is nonlinear. Second, the exponential crossover cannot generate all of the $2^{D}$ possible trial vectors because of its sequential nature. Therefore, replacing the binomial crossover by the exponential crossover is not only ineffective, but it can also degrade the performance of BetaCOBL.

\subsubsection{Applying Multiple Exponential Crossover}

\begin{figure}[h]
\includegraphics[width=\linewidth]{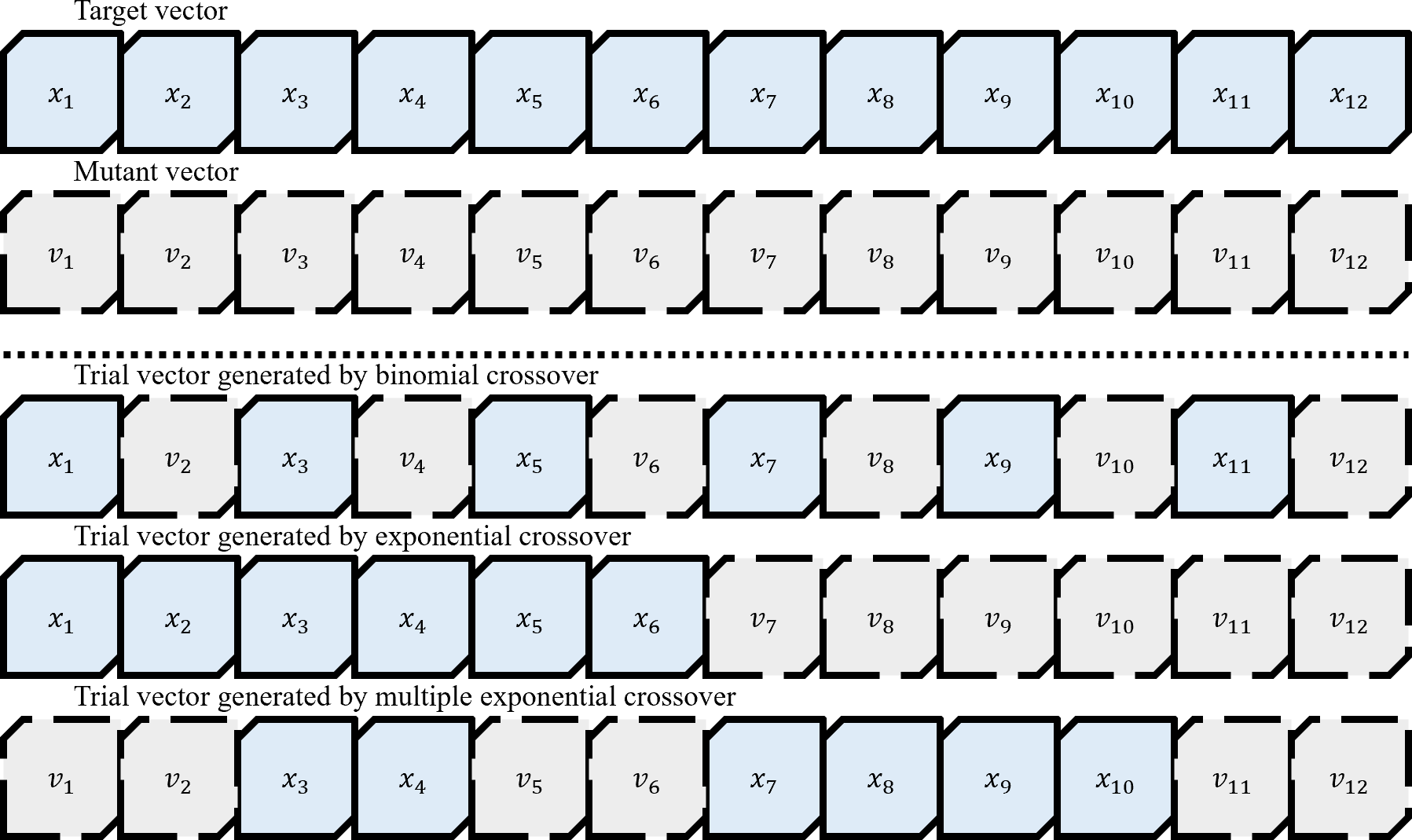}
\caption{Behavior of three crossover operators}
\centering
\label{fig:behaviorOfThreeCrossoverOperators}
\end{figure}

To improve the performance on inseparable problems, we employed the multiple exponential crossover \cite{MER} in the partial dimensional change scheme. The multiple exponential crossover is a semi-consecutive crossover operator that divides a trial vector into several components, and each component is a copy of the component at the location of either the target or the mutant vector \cite{MER}. Therefore, the multiple exponential crossover is the same as the exponential crossover that is being repeated. Fig. \ref{fig:behaviorOfThreeCrossoverOperators} shows the behavior of the binomial, exponential, and multiple exponential crossovers. In the proposed algorithm, the multiple exponential crossover calculates a partial opposite solution with a given pair of target vectors and complete opposite solution as follows. First, an element $n \in [1, D]$ is selected randomly. The four constants, $E_{m} = T \cdot CR$, $E_{s} = T \cdot (1 - CR)$, $CR_{m} = \frac{E_{m}}{E_{m} + 1}$, and $CR_{s} = \frac{E_{s}}{E_{s} + 1}$ are initialized where Em and Es stand for the approximate size of each component copied from the complete opposite solution and the target vector, respectively. Here, the length of the exchanged component $T$ is initialized at ten, as in \cite{MER}. Following this, the multiple exponential crossover calculates a partial opposite solution as follows:

\begin{enumerate}
\item Starting from the element $n$, a component of Bernoulli trials with $CR_{m}$ is calculated and copied from the complete opposite solution. \label{item:1}
\item Starting from the last failure element, the next component of Bernoulli trials with $CR_{s}$ is calculated and copied from the target vector. \label{item:2}
\item Repeat from Step \ref{item:1} until all of the elements are decided. \label{item:3}
\end{enumerate}

\noindent
The pseudo code of the multiple exponential crossover is presented in Algorithm \ref{alg:multipleExponentialCrossover}.

\begin{algorithm}
    \SetKwInOut{Input}{Input}
    \SetKwInOut{Output}{Output}

    \Input{Target vector $\vec{x}_{i,g}$, mutant vector $\vec{v}_{i,g}$, crossover rate $CR$, and length of exchanged components $T$}
    \Output{Trial vector $\vec{u}_{i,g}$}

    Select random integer $n$ within the range $[1,D]$\;
    $E_{m} = T \cdot CR$, $E_{s} = T \cdot (1 - CR)$\;
    $CR_{m} = \frac{E_{m}}{E_{m} + 1}$, $CR_{s} = \frac{E_{s}}{E_{s} + 1}$\;
    $L = 1$, $\textrm{Mutation\_Enable} = 1$\;
    \Repeat{$L \leq D$}{
        \eIf{$\textrm{Mutation\_Enable} == 1$}
        {
            \Repeat{$L \leq D$ and $rand_{i,j} \leq CR_{m}$}{
			    $u_{i,g}^{\langle n + L - 1 \rangle_{D}} = v_{i,g}^{\langle n + L - 1 \rangle_{D}}$\;
			    $L = L + 1$\;
            }
            $\textrm{Mutation\_Enable} = 0$\;
        }
        {
            \Repeat{$L \leq D$ and $rand_{i,j} \leq CR_{s}$}{
                $u_{i,g}^{\langle n + L - 1 \rangle_{D}} = x_{i,g}^{\langle n + L - 1 \rangle_{D}}$\;
			    $L = L + 1$\;
            }
            $\textrm{Mutation\_Enable} = 1$\;
        }
    }
    \caption{Multiple Exponential Crossover}
    \label{alg:multipleExponentialCrossover}
\end{algorithm}

\subsubsection{Rationale of Employing Multiple Exponential Crossover}

As mentioned in Section \ref{sec:Problem2}, it is of critical importance to preserve strongly dependent decision variables on inseparable problems when searching for satisfactory solutions. However, the exponential crossover is not an alternative to the binomial crossover as tuning the control parameter CR is difficult and it cannot generate all of the possible partial opposite solutions. Using a covariance matrix helps identify the inter-relations between decision variables, but it leads to a high computational cost. Therefore, we employed the multiple exponential crossover, which has the strengths of the exponential crossover but also retains the properties of the binomial crossover. With the replacement, the proposed algorithm can achieve better performance than BetaCOBL on inseparable problems by preserving the strongly dependent decision variables that are adjacent to each other.

\subsection{iBetaCODE}

iBetaCODE is the combination of DE and iBetaCOBL. As with other ODE variants, iBetaCOBL is executed in the initialization and iteration phases of iBetaCODE. In the initialization phase, iBetaCODE executes iBetaCOBL with the initialized individuals. In the iteration phase, iBetaCODE executes iBetaCOBL or the evolutionary operators of DE alternatively according to a predefined jumping rate $J_{r}$. If a random number generated according to the uniform distribution is lower than or equal to the jumping rate, iBetaCODE performs iBetaCOBL. Otherwise, iBetaCODE executes the evolutionary operators. Regarding the jumping rate, we set $J_{r} = 0.05$ in all the experiments in this paper, as in \cite{BetaCODE}. The entire pseudo code of iBetaCODE is presented in Algorithm \ref{alg:iBetaCODE}.

\begin{algorithm}
    \SetKwInOut{Input}{Input}
    \SetKwInOut{Output}{Output}

    \Input{Objective function $f(\vec{x})$, upper bound $\vec{x}_{max}$, lower bound $\vec{x}_{min}$, maximum number of function evaluations $NFEs_{max}$, scale factor $F$, crossover rate $CR$, population size $NP$, diversity threshold $DT$, and jumping rate, $J_{r}$}
    \Output{Best objective value $f(\vec{x}_{best})$}
    \tcc{Initialization phase}
    \For{$i = 0;\ i < NP;\ i = i + 1$}{
        \For{$j = 0;\ j < D;\ j = j + 1$}{
            $x_{i,0}^{j} = x_{min}^{j} + rand_{i}^{j} \cdot (x_{max}^{j} - x_{min}^{j})$\;
        }
    }
    $NFEs = NP, g = 1$\;
    Calculate $normDiv$ using equation (\ref{eqn:D_d'})\;
    \eIf{$normDiv > DT$}
    {
        $(\mu + \lambda)$ selection phase of iBetaCOBL (Algorithm \ref{alg:iBetaCOBL_1})\;
    }
    {
        $(\mu, \lambda)$ selection phase of iBetaCOBL (Algorithm \ref{alg:iBetaCOBL_2})\;
    }
    \tcc{Iteration phase}
    \While{None of termination criteria is satisfied}{
        \For{$i = 0;\ i < NP;\ i = i + 1$}{
            \eIf{$rand_{i} \leq J_{r}$}
            {
                Calculate $normDiv$ using equation (\ref{eqn:D_d'})\;
                \eIf{$normDiv > DT$}
                {
                    $(\mu + \lambda)$ selection phase of iBetaCOBL (Algorithm \ref{alg:iBetaCOBL_1})\;
                }
                {
                    $(\mu, \lambda)$ selection phase of iBetaCOBL (Algorithm \ref{alg:iBetaCOBL_2})\;
                }
            }
            {
                Select random three donor vectors $\vec{x}_{r_{1},g}$, $\vec{x}_{r_{2},g}$, $\vec{x}_{r_{3},g}$ where $r_{1} \neq r_{2} \neq r_{3} \neq i$\;
                Select random integer $j_{rand}$ within the range $[1,D]$\;
                \For{$j = 0;\ j < D;\ j = j + 1$}{
                    \eIf{$rand_{i}^{j} \leq CR$ or $j = j_{rand}$}
                    {
                        $u_{i,g}^{j} = x_{r_{1},g}^{j} + F \cdot (x_{r_{2},g}^{j} - x_{r_{3},g}^{j})$\;
                    }
                    {
                        $u_{i,g}^{j} = x_{i,g}^{j}$\;
                    }
                }
                \For{$i = 0;\ i < NP;\ i = i + 1$}{
                    \eIf{$f(\vec{u}_{i,g}) \leq f(\vec{x}_{i,g})$}
                    {
                        $\vec{x}_{i,g+1} = \vec{u}_{i,g}$\;
                    }
                    {
                        $\vec{x}_{i,g+1} = \vec{x}_{i,g}$\;
                    }
                }
                $NFEs = NFEs + NP$\;
            }
        }
        $g = g + 1$\;
    }
    \caption{iBetaCODE}
    \label{alg:iBetaCODE}
\end{algorithm}

\begin{algorithm}
    \SetKwInOut{Input}{Input}
    \SetKwInOut{Output}{Output}

    \Input{Population $\mathbf{P}_{g}$}
    \Output{Population $\mathbf{P}_{g}'$}
    
    \tcc{$(\mu + \lambda)$ selection}
    Set opposite population $\mathbf{OP}_{g} = (\mathbf{\breve{x}}_{1,g}, \mathbf{\breve{x}}_{2,g}, \cdots, \mathbf{\breve{x}}_{NP \cdot 2,g})$\;
    \For{$i = 0;\ i < NP;\ i = i + 1$}{
        \eIf{$rand_{i} \leq 0.5$}
        {
            Calculate spread using equation (\ref{eqn:concaveSpread})\;
            \For{$j = 0;\ j < D;\ j = j + 1$}{
                Calculate mode using equation (\ref{eqn:concaveMode})\;
                Calculate peak using equation (\ref{eqn:peak})\;
                Calculate alpha using equation (\ref{eqn:alpha})\;
                Calculate beta using equation (\ref{eqn:beta})\;
                $t_{i,g}^{j} = (x_{max}^{j} - x_{min}^{j}) \cdot Beta(\alpha, \beta) + x_{min}^{j}$\;
            }
        }
        {
            Calculate spread using equation (\ref{eqn:convexSpread})\;
            \For{$j = 0;\ j < D;\ j = j + 1$}{
                Calculate mode using equation (\ref{eqn:convexMode})\;
                Calculate peak using equation (\ref{eqn:peak})\;
                Calculate alpha using equation (\ref{eqn:alpha})\;
                Calculate beta using equation (\ref{eqn:beta})\;
                $t_{i,g}^{j} = (x_{max}^{j} - x_{min}^{j}) \cdot Beta(\alpha, \beta) + x_{min}^{j}$\;
            }
        }
        Calculate a partial opposite solution $\vec{\breve{x}}_{i,g}$ using Algorithm \ref{alg:multipleExponentialCrossover} with $\vec{t}_{i,g}$, $\vec{x}_{i,g}$, and $CR = 0.1$\;
        Calculate a partial opposite solution $\vec{\breve{x}}_{i+NP,g}$ using Algorithm \ref{alg:multipleExponentialCrossover} with $\vec{t}_{i,g}$, $\vec{x}_{i,g}$, and $CR = 0.9$\;
    }
    Merge original and opposite populations $\mathbf{P}_{g} + \mathbf{OP}_{g}$\;
    Select $NP$ best individuals $\mathbf{P}_{g}'$ from merged population $\mathbf{P}_{g} + \mathbf{OP}_{g}$\;
    $NFEs = NFEs + (NP \cdot 2)$\;
    
    \caption{$(\mu + \lambda)$ Selection Phase of iBetaCOBL}
    \label{alg:iBetaCOBL_1}
\end{algorithm}

\begin{algorithm}
    \SetKwInOut{Input}{Input}
    \SetKwInOut{Output}{Output}

    \Input{Population $\mathbf{P}_{g}$}
    \Output{Population $\mathbf{P}_{g}'$}
    
    \tcc{$(\mu, \lambda)$ selection}
    Sort population $\mathbf{P}_{g}$\;
    \For{$i = \frac{NP}{2};\ i < NP;\ i = i + 1$}{
        \eIf{$rand_{i} \leq 0.5$}
        {
            Calculate spread using equation (\ref{eqn:concaveSpread})\;
            \For{$j = 0;\ j < D;\ j = j + 1$}{
                Calculate mode using equation (\ref{eqn:concaveMode})\;
                Calculate peak using equation (\ref{eqn:peak})\;
                Calculate alpha using equation (\ref{eqn:alpha})\;
                Calculate beta using equation (\ref{eqn:beta})\;
                $t_{i,g}^{j} = (x_{max}^{j} - x_{min}^{j}) \cdot Beta(\alpha, \beta) + x_{min}^{j}$\;
            }
        }
        {
            Calculate spread using equation (\ref{eqn:convexSpread})\;
            \For{$j = 0;\ j < D;\ j = j + 1$}{
                Calculate mode using equation (\ref{eqn:convexMode})\;
                Calculate peak using equation (\ref{eqn:peak})\;
                Calculate alpha using equation (\ref{eqn:alpha})\;
                Calculate beta using equation (\ref{eqn:beta})\;
                $t_{i,g}^{j} = (x_{max}^{j} - x_{min}^{j}) \cdot Beta(\alpha, \beta) + x_{min}^{j}$\;
            }
        }
        Calculate a partial opposite solution $\vec{\breve{x}}_{1,g}$ using Algorithm \ref{alg:multipleExponentialCrossover} with $\vec{t}_{i,g}$, $\vec{x}_{i,g}$, and $CR = 0.1$\;
        Calculate a partial opposite solution $\vec{\breve{x}}_{2,g}$ using Algorithm \ref{alg:multipleExponentialCrossover} with $\vec{t}_{i,g}$, $\vec{x}_{i,g}$, and $CR = 0.9$\;
        \eIf{$f(\vec{\breve{x}}_{1,g}) \leq f(\vec{\breve{x}}_{2,g})$}
        {
            \If{$f(\vec{\breve{x}}_{1,g}) \leq f(\vec{x}_{i,g})$}
            {
                $\vec{x}_{i,g} = \vec{\breve{x}}_{1,g}$\;
            }
        }
        {
            \If{$f(\vec{\breve{x}}_{2,g}) \leq f(\vec{x}_{i,g})$}
            {
                $\vec{x}_{i,g} = \vec{\breve{x}}_{2,g}$\;
            }
        }
    }
    $NFEs = NFEs + NP$\;
    
    \caption{$(\mu, \lambda)$ Selection Phase of iBetaCOBL}
    \label{alg:iBetaCOBL_2}
\end{algorithm}


\section{Experimental Setup}
\label{sec:ExperimentalSetup}

All the experiments were conducted on Windows 10 Pro 64 bit of a PC with AMD Ryzen Threadripper 2990WX @ 3.0GHz. All the test algorithms were implemented in the C++ programming language with Visual Studio 2019 64 bit.

\subsection{Test Functions}

We utilized a set of 58 test functions for demonstrating the performance of the proposed algorithm. There are four well-known test suites on single objective bound constrained real-parameter numerical optimization, such as the CEC 2005, 2013, 2014, and 2017 test suites. We chose the CEC 2013 and 2017 test suites because the former is a directly improved version of the CEC 2005 test suite, while the latter is a directly improved version of the CEC 2014 test suite. In the CEC 2013 test suite, there are five unimodal functions ($F_{1}$-$F_{5}$), fifteen simple multimodal functions ($F_{6}$-$F_{20}$), and eight composition functions ($F_{21}$-$F_{28}$). In the CEC 2017 test suite, there are three unimodal functions ($F_{1}$-$F_{3}$), seven simple multimodal functions ($F_{4}$-$F_{10}$), ten expanded multimodal functions ($F_{11}$-$F_{20}$), and ten hybrid composition functions ($F_{21}$-$F_{30}$). For more detail explanations of the CEC 2013 and 2017 test suites, please refer to the following technical reports \cite{CEC2013, CEC2017}.

The experimental settings, such as the number of runs, the maximum number of function evaluations, and the minimum and maximum bounds, are initialized in the same way as in \cite{CEC2013} and \cite{CEC2017}.

\subsection{Performance Metrics}

\subsubsection{Function Error Value}

We utilized function error value (FEV) to evaluate the accuracy of a test algorithm, which can be defined as follows.

\begin{equation}
\textrm{FEV} = f(\vec{x}_{best,g_{max}}) - f(\vec{x}_{*})
\end{equation}

\noindent
where $f(x)$ stands for an objective function to be minimized. Also, $\vec{x}_{best,g_{max}}$ is the best solution found by a test algorithm, and $\vec{x}_{*}$ is the global optimum of a given problem. The lower the value of FEV, the higher the accuracy of a test algorithm.

\subsubsection{Statistical Test}

To determine whether the difference in performance for two test algorithms is significant or not, we utilized the Wilcoxon rank-sum test with $\alpha = 0.05$ significance level \cite{statistical}. The symbols in this paper have the following meanings unless stated otherwise.

\begin{enumerate}
\item +: The corresponding algorithm finds significantly better solutions than the proposed algorithm. \label{item:1}
\item =: The performance difference between the proposed algorithm and the corresponding algorithm is not statistically significant. \label{item:2}
\item -: The corresponding algorithm finds significantly worse solutions than the proposed algorithm. \label{item:3}
\end{enumerate}

Also, to determine whether the difference in performance for multiple test algorithms is significant or not, we utilized the Friedman test with Hochberg’s post hoc \cite{statistical}.


\section{Results and Comparisons}
\label{sec:ResultsAndComparisons}

\subsection{Comparison with Ten OBL Variants}
\label{sec:ResultsAndComparisons_TenODEVariants}

We performed experiments to evaluate the performance of iBetaCOBL and compared it to ten state-of-the-art OBL variants, namely: 1) OBL \cite{ODE}, 2) OBLTVJR \cite{ODETVJR}, 3) OBLPGJ \cite{ODEPGJ}, 4) QOBL \cite{QODE}, 5) QROBL \cite{QRODE}, 6) COOBL \cite{COODE}, 7) GOBL \cite{GODE}, 8) EOBL \cite{EODE}, 9) AGOBL \cite{AGODE}, and 10) BetaCOBL \cite{BetaCODE}. For a fair comparison, we used the same classical DE variant called DE/rand/1/bin; regarding the control parameters associated with the DE variant, we used the following values: $F = 0.5$, $CR = 0.9$, and $NP = 100$. Additionally, we used the values recommended by the authors of each paper for the remaining control parameters.

\subsubsection{Performance Evaluation on CEC 2013 Test Suite}

In this subsection, the performance evaluation results on the CEC 2013 test suite are presented. Twenty-eight benchmark problems from the CEC 2013 test suite are utilized to evaluate the performance of the test algorithms. Both 30-$D$ and 50-$D$ versions of the benchmark problems are tested.

Table \ref{tab:tenODEVariants_cec13_d30} shows the averages and standard deviations of the FEVs of each algorithm at 30 dimension, collected through 51 independent runs. As we can see from the table, the proposed algorithm has a clear edge over all the other OBL variants. More specifically, iBetaCOBL found more significantly accurate solutions than COOBL, OBL, OBLTVJR, QOBL, and QROBL on more than half of the test functions. In particular, iBetaCOBL considerably outperformed COOBL and QROBL on approximately four-fifths of the test functions. The second and third best algorithms are BetaCOBL and OBLPGJ, respectively. Compared with the original DE/rand/1/bin, DE/rand/1/bin assisted by iBetaCOBL considerably outperformed it on 12 test functions and underperformed it on 5 test functions. Compared with BetaCOBL, iBetaCOBL considerably outperformed it on 12 test functions and underperformed it on 6 test functions. In a word, DE/rand/1/bin assisted by iBetaCOBL secures an overall better performance than all the other OBL variants.

Also, Table \ref{tab:friedman_tenODEVariants_cec13_d30} shows the Friedman test with Hochberg's post hoc, which supports the experimental results in Table \ref{tab:tenODEVariants_cec13_d30} where iBetaCOBL ranked the first among the test algorithms, and the outperformance over COOBL, EOBL, OBL, OBLTVJR, and QROBL was statistically significant. In summary, the proposed algorithm is superior to the test algorithms on the CEC 2013 test suite at 30 dimension.

Additionally, we analyzed the performance evaluation results in Table \ref{tab:tenODEVariants_cec13_d30} based on the attributes of the test functions. The proposed algorithm achieved a similar optimization performance comparatively on the unimodal functions ($F_{1}$-$F_{5}$). However, it achieved a significantly better optimization performance in solving the multimodal ($F_{6}$-$F_{20}$) and composition functions ($F_{21}$-$F_{28}$). The results revealed that the proposed algorithm has a strong exploration property, and is thus capable of discovering more satisfactory solutions comparatively for more complex test functions.

We found similar tendencies at 50 dimension in Tables \ref{tab:tenODEVariants_cec13_d50} and \ref{tab:friedman_tenODEVariants_cec13_d50}. Compared with the experimental results at 30 dimension, the outperformance of the proposed algorithm is slightly larger at 50 dimension. For example, iBetaCOBL found more significantly accurate solutions on more than half the test functions compared with all the test algorithms except BetaCOBL. In particular, iBetaCOBL considerably outperformed COOBL, QOBL, and QROBL on approximately four-fifths of the test functions. Therefore, the proposed algorithm demonstrates that it can achieve better searchability than all the compared ones, including its predecessor BetaCOBL, particularly in the optimization for the multimodal and composition functions of the CEC 2013 test suite at both 30 and 50 dimensions.


\begin{table*}[htbp]
  \tiny
  \centering
  \caption{Averages and standard deviations of FEVs of DE/rand/1/bin with OBL variants on CEC 2013 test suite at 30-$D$.}
\begin{tabular}{cccccccc}
    \toprule
          & DE/rand/1/bin &       &       &       &       &       &  \\
    \cmidrule(l){2-2}
          & iBetaCOBL & Original & AGOBL & BetaCOBL & COOBL & EOBL  & GOBL \\
          & MEAN (STD DEV) & MEAN (STD DEV) & MEAN (STD DEV) & MEAN (STD DEV) & MEAN (STD DEV) & MEAN (STD DEV) & MEAN (STD DEV) \\
    \midrule
    F1    & 0.00E+00 (0.00E+00) & 0.00E+00 (0.00E+00) = & 0.00E+00 (0.00E+00) = & 0.00E+00 (0.00E+00) = & 0.00E+00 (0.00E+00) = & 0.00E+00 (0.00E+00) = & 0.00E+00 (0.00E+00) = \\
    F2    & 4.01E+05 (2.25E+05) & 3.71E+05 (2.96E+05) = & 4.77E+05 (2.96E+05) = & 4.86E+05 (3.34E+05) = & 1.50E+05 (9.22E+04) + & 4.55E+05 (2.79E+05) = & 5.95E+05 (4.86E+05) = \\
    F3    & 1.69E+03 (1.11E+04) & 7.30E-01 (4.09E+00) + & 2.75E+04 (1.95E+05) = & 1.74E+02 (9.14E+02) = & 1.42E+08 (9.67E+08) - & 5.18E+00 (2.77E+01) + & 3.90E+00 (1.60E+01) = \\
    F4    & 1.06E+03 (6.81E+02) & 9.68E+02 (4.64E+02) = & 1.13E+03 (6.27E+02) = & 1.38E+03 (5.49E+02) - & 1.61E+04 (3.04E+04) - & 2.87E+04 (1.07E+05) - & 1.05E+03 (4.58E+02) = \\
    F5    & 9.39E-14 (4.39E-14) & 8.72E-14 (4.88E-14) = & 8.94E-14 (4.74E-14) = & 9.16E-14 (4.57E-14) = & 7.82E-14 (5.34E-14) = & 9.16E-14 (4.57E-14) = & 7.60E-14 (5.43E-14) = \\
\bottomrule
    \end{tabular}%
  \label{tab:tenODEVariants_cec13_d30}%
\footnotesize
\\The symbols ``+/=/-" indicate that DE/rand/1/bin with a given OBL performed significantly better ($+$), not significantly better or worse ($=$), or significantly worse ($-$) compared to DE/rand/1/bin with iBetaCOBL using the Wilcoxon rank-sum test with $\alpha = 0.05$ significance level.
\end{table*}%

\begin{table*}[htbp]
  \tiny
  \centering
  \caption{Friedman test with Hochberg's post hoc for DE/rand/1/bin with OBL variants on CEC 2013 test suite at 30-$D$.}
    \begin{tabular}{cccccccccc}
    \toprule
          & DE/rand/1/bin &       &       &       &       &       &       &       &  \\
    \cmidrule(l){2-2}
          & Algorithm & Average ranking & z-value & p-value & Adj. p-value (Hochberg) & Sig.  &       & Test statistics &  \\
    \midrule
    1     & iBetaCOBL & 4.54  &       &       &       &       &       &       &  \\
    2     & Original & 5.63  & -1.13.E+00 & 2.58.E-01 & 7.75.E-01 & No    &       & N     & 28  \\
    3     & AGOBL & 5.88  & -1.39.E+00 & 1.65.E-01 & 8.23.E-01 & No    &       & Chi-Square & 50.25  \\
    4     & BetaCOBL & 4.80  & -2.78.E-01 & 7.81.E-01 & 7.81.E-01 & No    &       & df    & 11  \\
    5     & COOBL & 8.50  & -4.11.E+00 & 3.89.E-05 & 3.89.E-04 & Yes   &       & p-value & 5.65.E-07 \\
    6     & EOBL  & 7.38  & -2.95.E+00 & 3.21.E-03 & 2.25.E-02 & Yes   &       & Sig.  & Yes \\
    7     & GOBL  & 5.64  & -1.15.E+00 & 2.51.E-01 & 1.00.E+00 & No    &       &       &  \\
    8     & OBL   & 7.66  & -3.24.E+00 & 1.18.E-03 & 9.46.E-03 & Yes   &       &       &  \\
    9     & OBLPGJ & 5.00  & -4.82.E-01 & 6.30.E-01 & 1.26.E+00 & No    &       &       &  \\
    10    & OBLTVJR & 7.70  & -3.28.E+00 & 1.04.E-03 & 9.34.E-03 & Yes   &       &       &  \\
    11    & QOBL  & 6.57  & -2.11.E+00 & 3.46.E-02 & 2.08.E-01 & No    &       &       &  \\
    12    & QROBL & 8.71  & -4.34.E+00 & 1.45.E-05 & 1.59.E-04 & Yes   &       &       &  \\
    \bottomrule
    \end{tabular}%
  \label{tab:friedman_tenODEVariants_cec13_d30}%
\end{table*}%


\begin{table*}[htbp]
  \tiny
  \centering
  \caption{Averages and standard deviations of FEVs of DE/rand/1/bin with OBL variants on CEC 2013 test suite at 50-$D$.}
    \begin{tabular}{cccccccc}
    \toprule
          & DE/rand/1/bin &       &       &       &       &       &  \\
    \cmidrule(l){2-2}
          & iBetaCOBL & Original & AGOBL & BetaCOBL & COOBL & EOBL  & GOBL \\
          & MEAN (STD DEV) & MEAN (STD DEV) & MEAN (STD DEV) & MEAN (STD DEV) & MEAN (STD DEV) & MEAN (STD DEV) & MEAN (STD DEV) \\
    \midrule
    F1    & 4.45E-15 (3.18E-14) & 0.00E+00 (0.00E+00) = & 1.78E-14 (6.16E-14) = & 1.34E-14 (5.39E-14) = & 0.00E+00 (0.00E+00) = & 4.45E-15 (3.18E-14) = & 0.00E+00 (0.00E+00) = \\
    F2    & 1.48E+06 (6.36E+05) & 2.99E+06 (1.04E+06) - & 4.22E+06 (1.79E+06) - & 3.09E+06 (1.34E+06) - & 6.10E+05 (1.90E+05) + & 3.10E+06 (1.07E+06) - & 3.08E+06 (9.96E+05) - \\
    F3    & 5.97E+05 (1.29E+06) & 4.18E+05 (8.58E+05) = & 3.01E+05 (4.88E+05) = & 7.69E+05 (1.21E+06) = & 1.03E+09 (4.05E+09) - & 4.73E+05 (1.01E+06) = & 2.20E+05 (2.97E+05) = \\
    F4    & 6.22E+03 (1.59E+03) & 1.82E+04 (3.64E+03) - & 2.37E+04 (4.99E+03) - & 2.00E+04 (3.74E+03) - & 6.02E+04 (4.83E+04) - & 2.09E+04 (5.30E+03) - & 2.15E+04 (4.12E+03) - \\
    F5    & 1.12E-13 (1.60E-14) & 1.12E-13 (1.60E-14) = & 1.14E-13 (7.65E-29) = & 1.14E-13 (7.65E-29) = & 1.12E-13 (1.60E-14) = & 1.80E+02 (9.58E+02) = & 1.14E-13 (7.65E-29) = \\
\bottomrule
    \end{tabular}%
  \label{tab:tenODEVariants_cec13_d50}%
\footnotesize
\\The symbols ``+/=/-" indicate that DE/rand/1/bin with a given OBL performed significantly better ($+$), not significantly better or worse ($=$), or significantly worse ($-$) compared to DE/rand/1/bin with iBetaCOBL using the Wilcoxon rank-sum test with $\alpha = 0.05$ significance level.
\end{table*}%

\begin{table*}[htbp]
  \tiny
  \centering
  \caption{Friedman test with Hochberg's post hoc for DE/rand/1/bin with OBL variants on CEC 2013 test suite at 50-$D$.}
    \begin{tabular}{cccccccccc}
    \toprule
          & DE/rand/1/bin &       &       &       &       &       &       &       &  \\
    \cmidrule(l){2-2}
          & Algorithm & Average ranking & z-value & p-value & Adj. p-value (Hochberg) & Sig.  &       & Test statistics &  \\
    \midrule
    1     & iBetaCOBL & 3.80  &       &       &       &       &       &       &  \\
    2     & Original & 5.30  & -1.56.E+00 & 1.20.E-01 & 2.99.E-01 & No    &       & N     & 28  \\
    3     & AGOBL & 6.13  & -2.41.E+00 & 1.60.E-02 & 8.00.E-02 & No    &       & Chi-Square & 60.73  \\
    4     & BetaCOBL & 4.93  & -1.17.E+00 & 2.43.E-01 & 2.43.E-01 & No    &       & df    & 11  \\
    5     & COOBL & 8.36  & -4.73.E+00 & 2.30.E-06 & 2.30.E-05 & Yes   &       & p-value & 6.78.E-09 \\
    6     & EOBL  & 7.88  & -4.23.E+00 & 2.39.E-05 & 2.15.E-04 & Yes   &       & Sig.  & Yes \\
    7     & GOBL  & 5.30  & -1.56.E+00 & 1.20.E-01 & 2.99.E-01 & No    &       &       &  \\
    8     & OBL   & 7.84  & -4.19.E+00 & 2.81.E-05 & 2.25.E-04 & Yes   &       &       &  \\
    9     & OBLPGJ & 5.70  & -1.96.E+00 & 4.95.E-02 & 1.98.E-01 & No    &       &       &  \\
    10    & OBLTVJR & 7.05  & -3.37.E+00 & 7.44.E-04 & 5.21.E-03 & Yes   &       &       &  \\
    11    & QOBL  & 6.50  & -2.80.E+00 & 5.14.E-03 & 3.08.E-02 & Yes   &       &       &  \\
    12    & QROBL & 9.21  & -5.61.E+00 & 1.97.E-08 & 2.16.E-07 & Yes   &       &       &  \\
    \bottomrule
    \end{tabular}%
  \label{tab:friedman_tenODEVariants_cec13_d50}%
\end{table*}%



\begin{table*}[htbp]
  \tiny
  \centering
  \caption{Averages and standard deviations of FEVs of DE/rand/1/bin with OBL variants on CEC 2017 test suite at 30-$D$.}
    \begin{tabular}{cccccccc}
    \toprule
          & DE/rand/1/bin &       &       &       &       &       &  \\
    \cmidrule(l){2-2}
          & iBetaCOBL & Original & AGOBL & BetaCOBL & COOBL & EOBL  & GOBL \\
          & MEAN (STD DEV) & MEAN (STD DEV) & MEAN (STD DEV) & MEAN (STD DEV) & MEAN (STD DEV) & MEAN (STD DEV) & MEAN (STD DEV) \\
    \midrule
    F1    & 1.92E-14 (2.73E-14) & 5.57E-15 (7.00E-15) + & 3.37E-12 (6.47E-12) - & 7.38E-14 (1.26E-13) - & 4.59E-13 (2.62E-12) + & 2.54E-14 (3.71E-14) = & 2.95E-14 (5.06E-14) - \\
    F2    & 7.91E+09 (5.21E+10) & 4.35E+07 (1.98E+08) = & 1.26E+08 (6.21E+08) - & 5.93E+07 (4.02E+08) = & 1.48E+16 (1.06E+17) = & 1.79E+10 (1.16E+11) = & 1.48E+10 (1.05E+11) = \\
    F3    & 1.55E+00 (2.06E+00) & 1.35E+01 (2.15E+01) - & 4.06E+01 (3.87E+01) - & 5.91E+01 (7.09E+01) - & 1.03E+04 (3.77E+04) + & 3.82E+03 (2.69E+04) - & 2.94E+01 (3.62E+01) - \\
    F4    & 5.72E+01 (1.09E+01) & 5.71E+01 (1.18E+01) = & 5.77E+01 (8.30E+00) = & 5.55E+01 (1.43E+01) = & 4.30E+01 (2.86E+01) = & 5.70E+01 (1.14E+01) = & 5.72E+01 (1.18E+01) = \\
    F5    & 3.17E+01 (1.49E+01) & 1.75E+02 (8.56E+00) - & 1.79E+02 (9.52E+00) - & 6.73E+01 (3.17E+01) - & 1.15E+02 (5.14E+01) - & 1.72E+02 (2.00E+01) - & 1.78E+02 (9.31E+00) - \\
\bottomrule
    \end{tabular}%
  \label{tab:tenODEVariants_cec17_d30}%
\footnotesize
\\The symbols ``+/=/-" indicate that DE/rand/1/bin with a given OBL performed significantly better ($+$), not significantly better or worse ($=$), or significantly worse ($-$) compared to DE/rand/1/bin with iBetaCOBL using the Wilcoxon rank-sum test with $\alpha = 0.05$ significance level.
\end{table*}%

\begin{table*}[htbp]
  \tiny
  \centering
  \caption{Friedman test with Hochberg's post hoc for DE/rand/1/bin with OBL variants on CEC 2017 test suite at 30-$D$.}
    \begin{tabular}{cccccccccc}
    \toprule
          & DE/rand/1/bin &       &       &       &       &       &       &       &  \\
    \cmidrule(l){2-2}
          & Algorithm & Average ranking & z-value & p-value & Adj. p-value (Hochberg) & Sig.  &       & Test statistics &  \\
    \midrule
    1     & iBetaCOBL & 3.62  &       &       &       &       &       &       &  \\
    2     & Original & 4.75  & -1.22.E+00 & 2.23.E-01 & 4.47.E-01 & No    &       & N     & 30  \\
    3     & AGOBL & 7.27  & -3.92.E+00 & 8.83.E-05 & 6.18.E-04 & Yes   &       & Chi-Square & 92.17  \\
    4     & BetaCOBL & 4.00  & -4.12.E-01 & 6.81.E-01 & 6.81.E-01 & No    &       & df    & 11  \\
    5     & COOBL & 8.45  & -5.19.E+00 & 2.08.E-07 & 1.87.E-06 & Yes   &       & p-value & 6.26.E-15 \\
    6     & EOBL  & 6.62  & -3.22.E+00 & 1.27.E-03 & 7.62.E-03 & Yes   &       & Sig.  & Yes \\
    7     & GOBL  & 6.33  & -2.92.E+00 & 3.52.E-03 & 1.76.E-02 & Yes   &       &       &  \\
    8     & OBL   & 8.65  & -5.41.E+00 & 6.42.E-08 & 6.42.E-07 & Yes   &       &       &  \\
    9     & OBLPGJ & 5.32  & -1.83.E+00 & 6.78.E-02 & 2.04.E-01 & No    &       &       &  \\
    10    & OBLTVJR & 8.40  & -5.14.E+00 & 2.77.E-07 & 2.22.E-06 & Yes   &       &       &  \\
    11    & QOBL  & 5.43  & -1.95.E+00 & 5.10.E-02 & 2.04.E-01 & No    &       &       &  \\
    12    & QROBL & 9.17  & -5.96.E+00 & 2.50.E-09 & 2.75.E-08 & Yes   &       &       &  \\
    \bottomrule
    \end{tabular}%
  \label{tab:friedman_tenODEVariants_cec17_d30}%
\end{table*}%


\begin{table*}[htbp]
  \tiny
  \centering
  \caption{Averages and standard deviations of FEVs of DE/rand/1/bin with OBL variants on CEC 2017 test suite at 50-$D$.}
    \begin{tabular}{cccccccc}
    \toprule
          & DE/rand/1/bin &       &       &       &       &       &  \\
    \cmidrule(l){2-2}
          & iBetaCOBL & Original & AGOBL & BetaCOBL & COOBL & EOBL  & GOBL \\
          & MEAN (STD DEV) & MEAN (STD DEV) & MEAN (STD DEV) & MEAN (STD DEV) & MEAN (STD DEV) & MEAN (STD DEV) & MEAN (STD DEV) \\
    \midrule
    F1    & 7.87E+02 (2.07E+03) & 4.95E+02 (1.83E+03) + & 6.02E+02 (1.13E+03) = & 3.46E+02 (9.08E+02) = & 3.88E+03 (5.32E+03) - & 1.70E+09 (1.22E+10) + & 1.16E+02 (3.00E+02) = \\
    F2    & 1.26E+24 (8.93E+24) & 2.98E+25 (1.26E+26) - & 3.16E+27 (1.92E+28) - & 4.96E+25 (2.79E+26) - & 3.03E+33 (1.27E+34) = & 1.25E+28 (8.88E+28) - & 2.19E+28 (1.55E+29) - \\
    F3    & 1.56E+04 (4.21E+03) & 5.79E+04 (1.22E+04) - & 7.41E+04 (1.18E+04) - & 5.59E+04 (1.16E+04) - & 1.71E+05 (1.32E+05) - & 6.24E+04 (1.13E+04) - & 6.31E+04 (9.72E+03) - \\
    F4    & 6.09E+01 (4.38E+01) & 8.65E+01 (4.97E+01) - & 6.74E+01 (5.01E+01) = & 6.81E+01 (4.61E+01) = & 8.38E+01 (4.83E+01) - & 7.64E+01 (4.46E+01) = & 6.25E+01 (4.21E+01) = \\
    F5    & 6.11E+01 (2.07E+01) & 3.53E+02 (1.36E+01) - & 3.45E+02 (4.07E+01) - & 2.05E+02 (8.08E+01) - & 2.27E+02 (9.08E+01) - & 3.51E+02 (1.82E+01) - & 3.53E+02 (1.24E+01) - \\
\bottomrule
    \end{tabular}%
  \label{tab:tenODEVariants_cec17_d50}%
\footnotesize
\\The symbols ``+/=/-" indicate that DE/rand/1/bin with a given OBL performed significantly better ($+$), not significantly better or worse ($=$), or significantly worse ($-$) compared to DE/rand/1/bin with iBetaCOBL using the Wilcoxon rank-sum test with $\alpha = 0.05$ significance level.
\end{table*}%

\begin{table*}[htbp]
  \tiny
  \centering
  \caption{Friedman test with Hochberg's post hoc for DE/rand/1/bin with OBL variants on CEC 2017 test suite at 50-$D$.}
    \begin{tabular}{cccccccccc}
    \toprule
          & DE/rand/1/bin &       &       &       &       &       &       &       &  \\
    \cmidrule(l){2-2}
          & Algorithm & Average ranking & z-value & p-value & Adj. p-value (Hochberg) & Sig.  &       & Test statistics &  \\
    \midrule
    1     & iBetaCOBL & 3.07  &       &       &       &       &       &       &  \\
    2     & Original & 6.33  & -3.51.E+00 & 4.50.E-04 & 2.70.E-03 & Yes   &       & N     & 30  \\
    3     & AGOBL & 6.15  & -3.31.E+00 & 9.26.E-04 & 4.63.E-03 & Yes   &       & Chi-Square & 98.69  \\
    4     & BetaCOBL & 3.67  & -6.45.E-01 & 5.19.E-01 & 5.19.E-01 & No    &       & df    & 11  \\
    5     & COOBL & 8.93  & -6.30.E+00 & 2.94.E-10 & 2.94.E-09 & Yes   &       & p-value & 3.24.E-16 \\
    6     & EOBL  & 7.63  & -4.91.E+00 & 9.32.E-07 & 6.53.E-06 & Yes   &       & Sig.  & Yes \\
    7     & GOBL  & 5.93  & -3.08.E+00 & 2.07.E-03 & 8.30.E-03 & Yes   &       &       &  \\
    8     & OBL   & 8.60  & -5.94.E+00 & 2.79.E-09 & 2.51.E-08 & Yes   &       &       &  \\
    9     & OBLPGJ & 5.65  & -2.77.E+00 & 5.52.E-03 & 1.66.E-02 & Yes   &       &       &  \\
    10    & OBLTVJR & 7.77  & -5.05.E+00 & 4.45.E-07 & 3.56.E-06 & Yes   &       &       &  \\
    11    & QOBL  & 5.17  & -2.26.E+00 & 2.41.E-02 & 4.82.E-02 & Yes   &       &       &  \\
    12    & QROBL & 9.10  & -6.48.E+00 & 9.12.E-11 & 1.00.E-09 & Yes   &       &       &  \\
    \bottomrule
    \end{tabular}%
  \label{tab:friedman_tenODEVariants_cec17_d50}%
\end{table*}%


\subsubsection{Performance Evaluation on CEC 2017 Test Suite}

The results of performance evaluations on the CEC 2017 test suite are summarized in this subsection. Thirty benchmark problems from the CEC 2017 test suite are utilized to evaluate the performance of the test algorithms. Both 30-$D$ and 50-$D$ versions of the benchmark problems are tested.

Table \ref{tab:tenODEVariants_cec17_d30} shows the FEV averages and standard deviations of each algorithm at 30 dimension, obtained from 51 independent runs. The proposed algorithm has a clear edge over all the other OBL variants, as we can see from the table. More specifically, on more than half of the test functions, iBetaCOBL found more statistically precise solutions than all the other OBL variants. In particular, on all the test functions, iBetaCOBL substantially outperformed QROBL. The second and third best algorithms are BetaCOBL and the original DE/rand/1/bin, respectively. Compared with the original DE/rand/1/bin, DE/rand/1/bin assisted by iBetaCOBL considerably outperformed it on 16 test functions and underperformed it on 3 test functions. Compared with BetaCOBL, iBetaCOBL considerably outperformed it on 12 test functions and underperformed it on 4 test functions. In a word, DE/rand/1/bin assisted by iBetaCOBL secures an overall better performance than all the other OBL variants.

Moreover, Table \ref{tab:friedman_tenODEVariants_cec17_d30} shows the Friedman test with Hochberg's post hoc, which supports the experimental results in Table \ref{tab:tenODEVariants_cec17_d30} where iBetaCOBL ranked the first among the test algorithms, and the outperformance over AGOBL, COOBL, EOBL, GOBL, OBL, OBLTVJR, and QROBL was statistically significant. In summary, the proposed algorithm is superior to the test algorithms on the CEC 2017 test suite at 30 dimension.

Furthermore, we analyzed the performance evaluation results in Table \ref{tab:tenODEVariants_cec17_d30} based on the attributes of the test functions. The proposed algorithm achieved a similar optimization performance comparatively on the unimodal functions ($F_{1}$-$F_{3}$). However, it achieved a significantly better optimization performance in solving the multimodal ($F_{4}$-$F_{10}$), expanded multimodal ($F_{11}$-$F_{20}$), and hybrid composition functions ($F_{21}$-$F_{30}$). The results revealed that the proposed algorithm has a strong exploration property, and is thus capable of discovering more satisfactory solutions comparatively for more complex test functions.

We found similar tendencies at 50 dimension in Tables \ref{tab:tenODEVariants_cec17_d50} and \ref{tab:friedman_tenODEVariants_cec17_d50}. Compared with the experimental results at 30 dimension, the outperformance of the proposed algorithm is slightly larger at 50 dimension. For example, iBetaCOBL found more significantly accurate solutions on more than half the test functions compared with all the test algorithms. In particular, iBetaCOBL considerably outperformed COOBL and QROBL on approximately four-fifths of the test functions. Therefore, the proposed algorithm demonstrates that it can achieve better searchability than all the compared ones, including its predecessor BetaCOBL, particularly in the optimization for the multimodal, expanded multimodal, and hybrid composition functions of the CEC 2017 test suite at both 30 and 50 dimensions.

\subsubsection{Algorithm Complexity}


\begin{table*}[htbp]
  \tiny
  \centering
  \caption{Algorithm complexity for DE/rand/1/bin with OBL variants on CEC 2013 test suite at 30-$D$.}
    \begin{tabular}{ccccccccccccccccc}
    \toprule
          &       &       & DE/rand/1/bin &       &       &       &       &       &       &       &       &       &       &       &       &  \\
    \cmidrule(l){4-5}
          &       &       & iBetaCOBL &       & Original &       & AGOBL &       & BetaCOBL &       & COOBL &       & EOBL  &       & GOBL  &  \\
    \cmidrule(l){4-5} \cmidrule(l){6-7} \cmidrule(l){8-9} \cmidrule(l){10-11} \cmidrule(l){12-13} \cmidrule(l){14-15} \cmidrule(l){16-17}
    d     & T0    & T1    & T2    & (T2-T1)/T0 & T2    & (T2-T1)/T0 & T2    & (T2-T1)/T0 & T2    & (T2-T1)/T0 & T2    & (T2-T1)/T0 & T2    & (T2-T1)/T0 & T2    & (T2-T1)/T0 \\
    \midrule
    30    & 66.0  & 669.0  & 1333.6  & 10.1  & 1259.8  & 9.0   & 1265.6  & 9.0   & 2626.4  & 29.7  & 1225.6  & 8.4   & 1270.2  & 9.1   & 1265.2  & 9.0  \\
    \midrule
    \midrule
          &       &       &       &       &       &       & OBL   &       & OBLPGJ &       & OBLTVJR &       & QOBL  &       & QROBL &  \\
    \cmidrule(l){8-9} \cmidrule(l){10-11} \cmidrule(l){12-13} \cmidrule(l){14-15} \cmidrule(l){16-17}
          &       &       &       &       &       &       & T2    & (T2-T1)/T0 & T2    & (T2-T1)/T0 & T2    & (T2-T1)/T0 & T2    & (T2-T1)/T0 & T2    & (T2-T1)/T0 \\
    \midrule
          &       &       &       &       &       &       & 1276.8 & 9.2   & 1253.6 & 8.9   & 1273.8 & 9.2   & 1274.0 & 9.2   & 1220.0 & 8.3 \\
    \bottomrule
    \end{tabular}%
  \label{tab:complexity_cec13_30}%
\end{table*}%

\begin{table*}[htbp]
  \tiny
  \centering
  \caption{Algorithm complexity for DE/rand/1/bin with OBL variants on CEC 2013 test suite at 50-$D$.}
    \begin{tabular}{ccccccccccccccccc}
    \toprule
          &       &       & DE/rand/1/bin &       &       &       &       &       &       &       &       &       &       &       &       &  \\
    \cmidrule(l){4-5}
          &       &       & iBetaCOBL &       & Original &       & AGOBL &       & BetaCOBL &       & COOBL &       & EOBL  &       & GOBL  &  \\
    \cmidrule(l){4-5} \cmidrule(l){6-7} \cmidrule(l){8-9} \cmidrule(l){10-11} \cmidrule(l){12-13} \cmidrule(l){14-15} \cmidrule(l){16-17}
    d     & T0    & T1    & T2    & (T2-T1)/T0 & T2    & (T2-T1)/T0 & T2    & (T2-T1)/T0 & T2    & (T2-T1)/T0 & T2    & (T2-T1)/T0 & T2    & (T2-T1)/T0 & T2    & (T2-T1)/T0 \\
    \midrule
    50    & 66.0  & 1088.0  & 3642.8  & 38.7  & 3433.2  & 35.5  & 3434.6  & 35.6  & 7045.6  & 90.3  & 3293.8  & 33.4  & 3446.0  & 35.7  & 3443.6  & 35.7  \\
    \midrule
    \midrule
          &       &       &       &       &       &       & OBL   &       & OBLPGJ &       & OBLTVJR &       & QOBL  &       & QROBL &  \\
    \cmidrule(l){8-9} \cmidrule(l){10-11} \cmidrule(l){12-13} \cmidrule(l){14-15} \cmidrule(l){16-17}
          &       &       &       &       &       &       & T2    & (T2-T1)/T0 & T2    & (T2-T1)/T0 & T2    & (T2-T1)/T0 & T2    & (T2-T1)/T0 & T2    & (T2-T1)/T0 \\
    \midrule
          &       &       &       &       &       &       & 3425.4 & 35.4  & 3410.8 & 35.2  & 3403.2 & 35.1  & 3460.8 & 36.0  & 3200.4 & 32.0 \\
    \bottomrule
    \end{tabular}%
  \label{tab:complexity_cec13_50}%
\end{table*}%

\begin{table*}[htbp]
  \tiny
  \centering
  \caption{Algorithm complexity for DE/rand/1/bin with OBL variants on CEC 2017 test suite at 30-$D$.}
    \begin{tabular}{ccccccccccccccccc}
    \toprule
          &       &       & DE/rand/1/bin &       &       &       &       &       &       &       &       &       &       &       &       &  \\
    \cmidrule(l){4-5}
          &       &       & iBetaCOBL &       & Original &       & AGOBL &       & BetaCOBL &       & COOBL &       & EOBL  &       & GOBL  &  \\
    \cmidrule(l){4-5} \cmidrule(l){6-7} \cmidrule(l){8-9} \cmidrule(l){10-11} \cmidrule(l){12-13} \cmidrule(l){14-15} \cmidrule(l){16-17}
    d     & T0    & T1    & T2    & (T2-T1)/T0 & T2    & (T2-T1)/T0 & T2    & (T2-T1)/T0 & T2    & (T2-T1)/T0 & T2    & (T2-T1)/T0 & T2    & (T2-T1)/T0 & T2    & (T2-T1)/T0 \\
    \midrule
    30    & 62.0  & 267.0  & 749.2  & 7.8   & 627.0  & 5.8   & 640.4  & 6.0   & 2354.2  & 33.7  & 656.0  & 6.3   & 643.0  & 6.1   & 640.8  & 6.0  \\
    \midrule
    \midrule
          &       &       &       &       &       &       & OBL   &       & OBLPGJ &       & OBLTVJR &       & QOBL  &       & QROBL &  \\
    \cmidrule(l){8-9} \cmidrule(l){10-11} \cmidrule(l){12-13} \cmidrule(l){14-15} \cmidrule(l){16-17}
          &       &       &       &       &       &       & T2    & (T2-T1)/T0 & T2    & (T2-T1)/T0 & T2    & (T2-T1)/T0 & T2    & (T2-T1)/T0 & T2    & (T2-T1)/T0 \\
    \midrule
          &       &       &       &       &       &       & 656.4 & 6.3   & 626.0 & 5.8   & 649.0 & 6.2   & 643.2 & 6.1   & 756.8 & 7.9 \\
    \bottomrule
    \end{tabular}%
  \label{tab:complexity_cec17_30}%
\end{table*}%

\begin{table*}[htbp]
  \tiny
  \centering
  \caption{Algorithm complexity for DE/rand/1/bin with OBL variants on CEC 2017 test suite at 50-$D$.}
    \begin{tabular}{ccccccccccccccccc}
    \toprule
          &       &       & DE/rand/1/bin &       &       &       &       &       &       &       &       &       &       &       &       &  \\
    \cmidrule(l){4-5}
          &       &       & iBetaCOBL &       & Original &       & AGOBL &       & BetaCOBL &       & COOBL &       & EOBL  &       & GOBL  &  \\
    \cmidrule(l){4-5} \cmidrule(l){6-7} \cmidrule(l){8-9} \cmidrule(l){10-11} \cmidrule(l){12-13} \cmidrule(l){14-15} \cmidrule(l){16-17}
    d     & T0    & T1    & T2    & (T2-T1)/T0 & T2    & (T2-T1)/T0 & T2    & (T2-T1)/T0 & T2    & (T2-T1)/T0 & T2    & (T2-T1)/T0 & T2    & (T2-T1)/T0 & T2    & (T2-T1)/T0 \\
    \midrule
    50    & 62.0  & 505.0  & 2157.0  & 26.6  & 1776.0  & 20.5  & 1813.2  & 21.1  & 6982.8  & 104.5  & 1830.4  & 21.4  & 1819.4  & 21.2  & 1806.8  & 21.0  \\
    \midrule
    \midrule
          &       &       &       &       &       &       & OBL   &       & OBLPGJ &       & OBLTVJR &       & QOBL  &       & QROBL &  \\
    \cmidrule(l){8-9} \cmidrule(l){10-11} \cmidrule(l){12-13} \cmidrule(l){14-15} \cmidrule(l){16-17}
          &       &       &       &       &       &       & T2    & (T2-T1)/T0 & T2    & (T2-T1)/T0 & T2    & (T2-T1)/T0 & T2    & (T2-T1)/T0 & T2    & (T2-T1)/T0 \\
    \midrule
          &       &       &       &       &       &       & 1796.6 & 20.8  & 1784.4 & 20.6  & 1788.0 & 20.7  & 1812.4 & 21.1  & 2142.8 & 26.4 \\
    \bottomrule
    \end{tabular}%
  \label{tab:complexity_cec17_50}%
\end{table*}%


\begin{table}[htbp]
  \tiny
  \centering
  \caption{Averages and standard deviations of FEVs of DE/rand/1/bin with BetaCOBL variants on CEC 2013 test suite at 30-$D$.}
    \begin{tabular}{cccc}
    \toprule
          & DE/rand/1/bin &       &  \\
    \cmidrule(l){2-2}
          & BetaCOBL & BetaCOBL\_linear1 & BetaCOBL\_linear2 \\
          & MEAN (STD DEV) & MEAN (STD DEV) & MEAN (STD DEV) \\
    \midrule
    F1    & 0.00E+00 (0.00E+00) & 0.00E+00 (0.00E+00) = & 0.00E+00 (0.00E+00) = \\
    F2    & 4.86E+05 (3.34E+05) & 4.46E+05 (2.74E+05) = & 3.85E+05 (2.28E+05) = \\
    F3    & 1.74E+02 (9.14E+02) & 4.92E-01 (2.26E+00) = & 3.59E+04 (2.56E+05) = \\
    F4    & 1.38E+03 (5.49E+02) & 1.33E+03 (5.80E+02) = & 1.54E+03 (5.79E+02) = \\
    F5    & 9.16E-14 (4.57E-14) & 9.39E-14 (4.39E-14) = & 8.72E-14 (4.88E-14) = \\
    F6    & 1.06E+01 (6.38E+00) & 1.05E+01 (5.47E+00) = & 1.01E+01 (5.19E+00) = \\
    F7    & 1.38E-01 (2.01E-01) & 1.66E-01 (2.17E-01) = & 1.61E-01 (2.24E-01) = \\
    F8    & 2.10E+01 (5.77E-02) & 2.10E+01 (5.42E-02) = & 2.10E+01 (5.22E-02) = \\
    F9    & 6.46E+00 (2.28E+00) & 6.64E+00 (2.29E+00) = & 6.87E+00 (2.38E+00) = \\
    F10   & 6.67E-03 (5.58E-03) & 6.76E-03 (5.61E-03) = & 7.15E-03 (6.54E-03) = \\
    F11   & 4.78E+01 (1.04E+01) & 4.70E+01 (1.13E+01) = & 4.59E+01 (8.82E+00) = \\
    F12   & 1.74E+02 (1.13E+01) & 1.72E+02 (1.31E+01) = & 1.75E+02 (1.27E+01) = \\
    F13   & 1.76E+02 (1.19E+01) & 1.80E+02 (1.00E+01) = & 1.76E+02 (1.20E+01) = \\
    F14   & 1.11E+03 (2.80E+02) & 1.09E+03 (2.74E+02) = & 1.07E+03 (2.45E+02) = \\
    F15   & 7.06E+03 (2.84E+02) & 7.06E+03 (2.39E+02) = & 6.90E+03 (6.26E+02) = \\
    F16   & 2.50E+00 (2.39E-01) & 2.42E+00 (2.56E-01) = & 2.42E+00 (2.61E-01) = \\
    F17   & 1.03E+02 (1.19E+01) & 1.01E+02 (8.90E+00) = & 1.01E+02 (1.09E+01) = \\
    F18   & 2.08E+02 (1.01E+01) & 2.10E+02 (7.69E+00) = & 2.10E+02 (1.13E+01) = \\
    F19   & 1.14E+01 (1.46E+00) & 1.17E+01 (1.24E+00) = & 1.16E+01 (1.42E+00) = \\
    F20   & 1.22E+01 (2.21E-01) & 1.22E+01 (3.09E-01) = & 1.21E+01 (2.49E-01) = \\
    F21   & 3.15E+02 (8.58E+01) & 2.91E+02 (8.87E+01) = & 2.97E+02 (8.57E+01) = \\
    F22   & 1.07E+03 (2.50E+02) & 1.04E+03 (2.69E+02) = & 1.07E+03 (2.60E+02) = \\
    F23   & 6.93E+03 (3.44E+02) & 6.98E+03 (3.02E+02) = & 6.86E+03 (4.26E+02) = \\
    F24   & 2.00E+02 (0.00E+00) & 2.00E+02 (0.00E+00) = & 2.00E+02 (0.00E+00) = \\
    F25   & 2.40E+02 (4.75E+00) & 2.39E+02 (4.10E+00) = & 2.40E+02 (5.01E+00) = \\
    F26   & 2.00E+02 (0.00E+00) & 2.00E+02 (0.00E+00) = & 2.00E+02 (0.00E+00) = \\
    F27   & 3.04E+02 (1.91E+01) & 3.02E+02 (2.04E+00) - & 3.10E+02 (3.47E+01) = \\
    F28   & 3.00E+02 (0.00E+00) & 3.00E+02 (0.00E+00) = & 3.00E+02 (0.00E+00) = \\
    \midrule
    +/=/- &       & 0/27/1 & 0/28/0 \\
    \bottomrule
    \end{tabular}%
  \label{tab:linearBetaCODEVariants_cec13_d30}%
\\The symbols ``+/=/-" indicate that DE/rand/1/bin with a given linear time diversity measure based BetaCOBL performed significantly better ($+$), not significantly better or worse ($=$), or significantly worse ($-$) compared to DE/rand/1/bin with BetaCOBL using the Wilcoxon rank-sum test with $\alpha = 0.05$ significance level.
\end{table}%

\begin{table}[htbp]
  \tiny
  \centering
  \caption{Averages and standard deviations of FEVs of DE/rand/1/bin with BetaCOBL variants on CEC 2017 test suite at 30-$D$.}
    \begin{tabular}{cccc}
    \toprule
          & DE/rand/1/bin &       &  \\
    \cmidrule(l){2-2}
          & BetaCOBL & BetaCOBL\_linear1 & BetaCOBL\_linear2 \\
          & MEAN (STD DEV) & MEAN (STD DEV) & MEAN (STD DEV) \\
    \midrule
    F1    & 7.38E-14 (1.26E-13) & 1.38E-13 (2.77E-13) = & 2.53E-13 (5.41E-13) - \\
    F2    & 5.93E+07 (4.02E+08) & 2.46E+08 (1.52E+09) = & 2.06E+08 (9.50E+08) = \\
    F3    & 5.91E+01 (7.09E+01) & 5.56E+01 (5.19E+01) = & 4.63E+01 (3.71E+01) = \\
    F4    & 5.55E+01 (1.43E+01) & 5.81E+01 (8.14E+00) = & 5.80E+01 (8.43E+00) = \\
    F5    & 6.73E+01 (3.17E+01) & 6.75E+01 (2.94E+01) = & 7.22E+01 (3.19E+01) = \\
    F6    & 1.26E-07 (1.37E-07) & 1.51E-07 (1.44E-07) = & 1.46E-07 (1.36E-07) = \\
    F7    & 1.75E+02 (2.39E+01) & 1.73E+02 (2.52E+01) = & 1.73E+02 (2.74E+01) = \\
    F8    & 6.97E+01 (3.28E+01) & 6.64E+01 (3.01E+01) = & 6.36E+01 (3.22E+01) = \\
    F9    & 0.00E+00 (0.00E+00) & 0.00E+00 (0.00E+00) = & 0.00E+00 (0.00E+00) = \\
    F10   & 3.25E+03 (6.98E+02) & 3.21E+03 (9.59E+02) = & 3.04E+03 (7.26E+02) = \\
    F11   & 2.04E+01 (2.05E+01) & 1.61E+01 (1.57E+01) = & 1.50E+01 (1.49E+01) = \\
    F12   & 9.07E+03 (6.79E+03) & 8.46E+03 (5.20E+03) = & 8.80E+03 (5.88E+03) = \\
    F13   & 8.53E+01 (8.72E+00) & 8.29E+01 (1.46E+01) = & 8.07E+01 (7.96E+00) + \\
    F14   & 1.38E+01 (6.54E+00) & 1.42E+01 (6.85E+00) = & 1.31E+01 (6.65E+00) = \\
    F15   & 9.59E+00 (4.53E+00) & 1.14E+01 (6.41E+00) = & 9.61E+00 (4.94E+00) = \\
    F16   & 5.30E+02 (2.59E+02) & 5.00E+02 (2.52E+02) = & 5.76E+02 (2.72E+02) = \\
    F17   & 9.29E+01 (7.85E+01) & 9.42E+01 (1.04E+02) = & 1.14E+02 (1.18E+02) = \\
    F18   & 2.63E+01 (3.47E+00) & 2.87E+01 (1.45E+01) = & 2.58E+01 (4.42E+00) = \\
    F19   & 7.45E+00 (1.74E+00) & 8.03E+00 (1.82E+00) = & 7.57E+00 (2.14E+00) = \\
    F20   & 1.43E+02 (1.17E+02) & 1.58E+02 (1.26E+02) = & 1.24E+02 (1.34E+02) = \\
    F21   & 2.53E+02 (3.09E+01) & 2.53E+02 (2.73E+01) = & 2.56E+02 (2.78E+01) = \\
    F22   & 1.00E+02 (0.00E+00) & 1.00E+02 (0.00E+00) = & 1.00E+02 (0.00E+00) = \\
    F23   & 3.94E+02 (4.31E+01) & 3.89E+02 (3.20E+01) = & 3.87E+02 (3.02E+01) = \\
    F24   & 5.18E+02 (6.13E+01) & 5.05E+02 (6.07E+01) = & 4.85E+02 (5.79E+01) + \\
    F25   & 3.87E+02 (0.00E+00) & 3.87E+02 (0.00E+00) = & 3.87E+02 (0.00E+00) = \\
    F26   & 1.24E+03 (3.57E+02) & 1.19E+03 (2.43E+02) = & 1.26E+03 (3.39E+02) = \\
    F27   & 4.82E+02 (1.20E+01) & 4.83E+02 (1.11E+01) = & 4.84E+02 (1.14E+01) = \\
    F28   & 3.32E+02 (4.98E+01) & 3.17E+02 (4.09E+01) = & 3.17E+02 (3.99E+01) = \\
    F29   & 4.75E+02 (4.56E+01) & 4.76E+02 (3.42E+01) = & 4.74E+02 (4.32E+01) = \\
    F30   & 2.02E+03 (5.26E+01) & 2.01E+03 (4.85E+01) = & 2.01E+03 (4.50E+01) = \\
    \midrule
    +/=/- &       & 0/30/0 & 2/27/1 \\
    \bottomrule
    \end{tabular}%
  \label{tab:linearBetaCODEVariants_cec17_d30}%
\\The symbols ``+/=/-" indicate that DE/rand/1/bin with a given linear time diversity measure based BetaCOBL performed significantly better ($+$), not significantly better or worse ($=$), or significantly worse ($-$) compared to DE/rand/1/bin with BetaCOBL using the Wilcoxon rank-sum test with $\alpha = 0.05$ significance level.
\end{table}%


In this subsection, the algorithm complexity results on the CEC 2013 and 2017 test suites are presented. Table \ref{tab:complexity_cec13_30} and \ref{tab:complexity_cec13_50} show the experimentally estimated algorithm complexity of each algorithm according to the CEC 2013 test suite at 30 and 50-dimensions, respectively. As we can see from the tables, the proposed algorithm consumed approximately similar or slightly higher computational time in comparison with the other OBL variants except for the original BetaCOBL. The original BetaCOBL consumed approximately three times higher computational time than the proposed algorithm at both of the dimensions. We found similar tendencies for the results on the CEC 2017 test suite in Tables \ref{tab:complexity_cec17_30} and \ref{tab:complexity_cec17_50}. As a result, the proposed algorithm consumed significantly less computational time than the original BetaCOBL even though it found significantly better solutions.

\subsection{BetaCOBL with Linear Time Diversity Measures}

To reduce the computational cost, we employed the linear time diversity measure $D_{d}$ in the selection switching scheme instead of using the power mean-based diversity measure $D_{h}$. However, replacing the power mean by the linear time may lead to performance issues. To investigate the impact of replacing the diversity measure, we compared the original BetaCOBL with two linear time BetaCOBL variants. As in the previous experiments, we used the same classical DE variant called DE/rand/1/bin, and for the control parameters associated with the DE variant, with the following values: $F = 0.5$, $CR = 0.9$, and $NP = 100$. Additionally, we used $DT = 1e-6$ for the diversity threshold and $J_{r} = 0.05$ for the jumping rate.

Tables \ref{tab:linearBetaCODEVariants_cec13_d30} and \ref{tab:linearBetaCODEVariants_cec17_d30} show the averages and standard deviations of the FEVs of each algorithm on the CEC 2013 and 2017 test suites, respectively. In the tables, BetaCOBL\_linear1 and BetaCOBL\_linear2 stand for BetaCOBL using the diversity measure $D_{v}$ and $D_{d}$, respectively. As we can see from the tables, the performance difference between the original BetaCOBL and the linear time BetaCOBL variants is negligible. On the CEC 2013 test suite, the original BetaCOBL outperformed BetaCOBL\_linear1 and BetaCOBL\_linear2 on one and zero test functions only, respectively. On the CEC 2017 test suite, the original BetaCOBL outperformed BetaCOBL\_linear1 and BetaCOBL\_linear2 on zero and one test functions only, respectively. However, BetaCOBL\_linear2 outperformed the original BetaCOBL on two test functions. The experimental results reveal that replacing the power mean-based diversity measure with any of the two linear time diversity measures does not adversely impact the performance of BetaCOBL. Notably, BetaCOBL\_linear2 was slightly better than the original BetaCOBL on the CEC 2017 test suite. Also, the diversity measure $D_{d}$ has been mathematically proven to discourage the collocation of individuals with a larger than the diversity measure $D_{v}$, as we explained in Section \ref{sec:ProposedAlgorithm_ModifiedSelectionSwitchingScheme_3}. Therefore, we chose the diversity measure $D_{d}$ for the proposed algorithm.

\section{Performance Enhancement of DE Variants}
\label{sec:PerformanceEnhancementOfDEVariants}

In the previous section, BetaCOBL has proven to be effective in improving a classical DE variant. We investigated further to check the compatibility of the proposed algorithm with two state-of-the-art DE variants, EDEV \cite{EDEV} and LSHADE-RSP \cite{LSHADE-RSP}.

EDEV is a multi-population-based DE variant, which consists of three DE variants, JADE \cite{JADE}, CoDE \cite{CoDE}, and EPSDE \cite{EPSDE}. EDEV uses a larger reward and three equally smaller populations. Each smaller population uses a distinct DE variant, and the larger reward population uses the best DE variant determined by comparing the success rate of each smaller population for every predefined number of generations. EDEV outperformed AEPD-JADE \cite{AEPD-JADE}, DE-VNS \cite{DE-VNS}, rank-jDE \cite{rank-jDE}, sinDE \cite{sinDE}, and MPEDE \cite{MPEDE}. LSHADE-RSP is an improved L-SHADE that uses a fast trial vector generation strategy, an external archive that stores discarded individuals, a historical memory-based adaptive parameter control, and a linear population size reduction. L-SHADE \cite{L-SHADE} secured the first rank on the CEC 2014 competition on numerical optimization and is basis of a set of powerful DE variants, such as iL-SHADE \cite{brest2016shade}, jSO \cite{brest2017single}, LSHADE-EpSin \cite{awad2016ensemble}, L-$conv$SHADE \cite{awad2017ensemble}, LSHADE-cnEpSin \cite{awad2017novel}, EsDE$_{\text{r}}$-NR \cite{awad2018ensemble}, LSHADE-SPACMA \cite{mohamed2017lshade}, LSHADE-RSP \cite{LSHADE-RSP}, and mL-SHADE \cite{yeh2019modified}. LSHADE-RSP uses a rank-based selective pressure to establish a balance between exploration and exploitation, which secured the second rank on the CEC 2018 competition on numerical optimization.

We used $DT = 1e-6$ for the diversity threshold and $J_{r} = 0.05$ for the jumping rate. Note that LSAHDE-RSP has a large population size at the beginning of an optimization process. Thus, all the OBL variants may be ineffective in the early stage of the optimization process. Therefore we modified that LSHADE-RSP starts to run OBL when it reaches three-fourths of the maximum number of function evaluations, which is the late stage of the optimization process. LSHADE-RSP has a small population size in the late stage of the optimization process. Note that the modification is for LSHADE-RSP, and EDEV starts to run OBL at the beginning of the optimization process.

\subsection{Performance Enhancement of EDEV Algorithm}

\begin{table*}[htbp]
  \tiny
  \centering
  \caption{Averages and standard deviations of FEVs of EDEV with OBL variants on CEC 2013 test suite at 50-$D$.}
\begin{tabular}{cccccccc}
    \toprule
          & EDEV &       &       &       &       &       &  \\
    \cmidrule(l){2-2}
          & iBetaCOBL & Original & AGOBL & BetaCOBL & COOBL & EOBL  & GOBL \\
          & MEAN (STD DEV) & MEAN (STD DEV) & MEAN (STD DEV) & MEAN (STD DEV) & MEAN (STD DEV) & MEAN (STD DEV) & MEAN (STD DEV) \\
    \midrule
    F1    & 0.00E+00 (0.00E+00) & 0.00E+00 (0.00E+00) = & 0.00E+00 (0.00E+00) = & 0.00E+00 (0.00E+00) = & 0.00E+00 (0.00E+00) = & 0.00E+00 (0.00E+00) = & 0.00E+00 (0.00E+00) = \\
    F2    & 4.82E+04 (2.26E+04) & 3.79E+04 (1.73E+04) + & 5.63E+04 (2.71E+04) = & 4.58E+04 (1.91E+04) = & 5.19E+04 (2.33E+04) = & 4.17E+04 (1.86E+04) = & 3.78E+04 (2.22E+04) + \\
    F3    & 5.11E+06 (1.92E+07) & 2.81E+06 (4.62E+06) = & 1.77E+06 (3.04E+06) = & 1.21E+06 (1.96E+06) = & 1.52E+08 (1.04E+09) - & 1.49E+06 (2.72E+06) = & 3.88E+06 (8.49E+06) = \\
    F4    & 1.40E+03 (6.17E+03) & 3.90E+03 (1.12E+04) + & 7.23E+03 (2.01E+04) + & 1.10E+03 (7.83E+03) = & 3.75E+04 (6.34E+04) - & 4.76E+03 (1.50E+04) + & 2.25E+03 (9.80E+03) + \\
    F5    & 1.05E-13 (3.10E-14) & 1.12E-13 (1.60E-14) = & 1.12E-13 (1.60E-14) = & 1.14E-13 (7.65E-29) = & 1.14E-13 (7.65E-29) = & 2.67E+03 (1.33E+04) = & 1.10E-13 (2.23E-14) = \\
\bottomrule
    \end{tabular}%
  \label{tab:EDEV_wilcoxon_cec13}%
\footnotesize
\\The symbols ``+/=/-" indicate that EDEV with a given OBL performed significantly better ($+$), not significantly better or worse ($=$), or significantly worse ($-$) compared to EDEV with iBetaCOBL using the Wilcoxon rank-sum test with $\alpha = 0.05$ significance level.
\end{table*}%

\begin{table*}[htbp]
  \tiny
  \centering
  \caption{Friedman test with Hochberg's post hoc for EDEV with OBL variants on CEC 2013 test suite at 50-$D$.}
    \begin{tabular}{cccccccccc}
    \toprule
          & EDEV &       &       &       &       &       &       &       &  \\
    \cmidrule(l){2-2}
          & Algorithm & Average ranking & z-value & p-value & Adj. p-value (Hochberg) & Sig.  &       & Test statistics &  \\
    \midrule
    1     & iBetaCOBL & 3.50  &       &       &       &       &       &       &  \\
    2     & Original & 5.09  & -1.65.E+00 & 9.91.E-02 & 1.98.E-01 & No    &       & N     & 28  \\
    3     & AGOBL & 6.38  & -2.98.E+00 & 2.85.E-03 & 1.71.E-02 & Yes   &       & Chi-Square & 87.26  \\
    4     & BetaCOBL & 4.82  & -1.37.E+00 & 1.70.E-01 & 1.70.E-01 & No    &       & df    & 11  \\
    5     & COOBL & 10.63  & -7.39.E+00 & 1.43.E-13 & 1.57.E-12 & Yes   &       & p-value & 5.74.E-14 \\
    6     & EOBL  & 7.59  & -4.24.E+00 & 2.20.E-05 & 1.76.E-04 & Yes   &       & Sig.  & Yes \\
    7     & GOBL  & 6.00  & -2.59.E+00 & 9.48.E-03 & 4.74.E-02 & Yes   &       &       &  \\
    8     & OBL   & 7.68  & -4.34.E+00 & 1.45.E-05 & 1.30.E-04 & Yes   &       &       &  \\
    9     & OBLPGJ & 5.41  & -1.98.E+00 & 4.74.E-02 & 1.66.E-01 & No    &       &       &  \\
    10    & OBLTVJR & 6.93  & -3.56.E+00 & 3.74.E-04 & 2.62.E-03 & Yes   &       &       &  \\
    11    & QOBL  & 5.41  & -1.98.E+00 & 4.74.E-02 & 1.66.E-01 & No    &       &       &  \\
    12    & QROBL & 8.57  & -5.26.E+00 & 1.42.E-07 & 1.42.E-06 & Yes   &       &       &  \\
    \bottomrule
    \end{tabular}%
  \label{tab:EDEV_posthoc_cec13}%
\end{table*}%

\begin{table*}[htbp]
  \tiny
  \centering
  \caption{Averages and standard deviations of FEVs of EDEV with OBL variants on CEC 2017 test suite at 50-$D$.}
    \begin{tabular}{cccccccc}
    \toprule
          & EDEV &       &       &       &       &       &  \\
    \cmidrule(l){2-2}
          & iBetaCOBL & Original & AGOBL & BetaCOBL & COOBL & EOBL  & GOBL \\
          & MEAN (STD DEV) & MEAN (STD DEV) & MEAN (STD DEV) & MEAN (STD DEV) & MEAN (STD DEV) & MEAN (STD DEV) & MEAN (STD DEV) \\
    \midrule
    F1    & 5.32E-14 (1.30E-13) & 3.48E-13 (1.87E-12) = & 4.23E-13 (2.52E-12) = & 4.00E-11 (2.74E-10) = & 5.84E-11 (2.51E-10) = & 2.75E-12 (1.56E-11) = & 5.14E-13 (2.44E-12) = \\
    F2    & 1.23E+20 (8.78E+20) & 1.48E+25 (7.91E+25) = & 1.06E+33 (7.56E+33) = & 1.77E+26 (1.21E+27) = & 8.80E+01 (4.91E+02) - & 6.29E+24 (4.47E+25) = & 8.05E+24 (3.10E+25) = \\
    F3    & 7.61E+03 (1.74E+04) & 1.70E+04 (3.88E+04) = & 3.04E+04 (5.14E+04) = & 1.84E+04 (4.03E+04) = & 6.96E+04 (1.12E+05) - & 2.16E+04 (4.34E+04) = & 1.24E+04 (3.54E+04) = \\
    F4    & 4.64E+01 (4.54E+01) & 6.29E+01 (4.40E+01) - & 5.68E+01 (4.89E+01) = & 5.67E+01 (4.97E+01) = & 4.66E+01 (4.71E+01) = & 2.13E+03 (1.48E+04) = & 5.45E+01 (4.67E+01) = \\
    F5    & 5.47E+01 (1.18E+01) & 6.44E+01 (8.68E+00) - & 6.78E+01 (1.29E+01) - & 5.91E+01 (9.66E+00) - & 1.82E+02 (9.61E+01) - & 6.21E+01 (1.08E+01) - & 6.26E+01 (9.45E+00) - \\
\bottomrule
    \end{tabular}%
  \label{tab:EDEV_wilcoxon_cec17}%
\footnotesize
\\The symbols ``+/=/-" indicate that EDEV with a given OBL performed significantly better ($+$), not significantly better or worse ($=$), or significantly worse ($-$) compared to EDEV with iBetaCOBL using the Wilcoxon rank-sum test with $\alpha = 0.05$ significance level.
\end{table*}%

\begin{table*}[htbp]
  \tiny
  \centering
  \caption{Friedman test with Hochberg's post hoc for EDEV with OBL variants on CEC 2017 test suite at 50-$D$.}
    \begin{tabular}{cccccccccc}
    \toprule
          & EDEV &       &       &       &       &       &       &       &  \\
    \cmidrule(l){2-2}
          & Algorithm & Average ranking & z-value & p-value & Adj. p-value (Hochberg) & Sig.  &       & Test statistics &  \\
    \midrule
    1     & iBetaCOBL & 2.87  &       &       &       &       &       &       &  \\
    2     & Original & 5.40  & -2.72.E+00 & 6.50.E-03 & 1.95.E-02 & Yes   &       & N     & 30  \\
    3     & AGOBL & 7.03  & -4.48.E+00 & 7.62.E-06 & 4.57.E-05 & Yes   &       & Chi-Square & 112.85  \\
    4     & BetaCOBL & 4.13  & -1.36.E+00 & 1.74.E-01 & 1.74.E-01 & No    &       & df    & 11  \\
    5     & COOBL & 9.93  & -7.59.E+00 & 3.18.E-14 & 3.50.E-13 & Yes   &       & p-value & 4.94.E-19 \\
    6     & EOBL  & 7.53  & -5.01.E+00 & 5.36.E-07 & 3.75.E-06 & Yes   &       & Sig.  & Yes \\
    7     & GOBL  & 5.77  & -3.12.E+00 & 1.84.E-03 & 9.19.E-03 & Yes   &       &       &  \\
    8     & OBL   & 8.52  & -6.07.E+00 & 1.29.E-09 & 1.16.E-08 & Yes   &       &       &  \\
    9     & OBLPGJ & 5.57  & -2.90.E+00 & 3.73.E-03 & 1.49.E-02 & Yes   &       &       &  \\
    10    & OBLTVJR & 8.07  & -5.59.E+00 & 2.33.E-08 & 1.86.E-07 & Yes   &       &       &  \\
    11    & QOBL  & 4.62  & -1.88.E+00 & 6.01.E-02 & 1.20.E-01 & No    &       &       &  \\
    12    & QROBL & 8.57  & -6.12.E+00 & 9.20.E-10 & 9.20.E-09 & Yes   &       &       &  \\
    \bottomrule
    \end{tabular}%
  \label{tab:EDEV_posthoc_cec17}%
\end{table*}%

\begin{table*}[htbp]
  \tiny
  \centering
  \caption{Algorithm complexity for EDEV with OBL variants on CEC 2013 test suite at 50-$D$.}
    \begin{tabular}{ccccccccccccccccc}
    \toprule
          &       &       & EDEV &       &       &       &       &       &       &       &       &       &       &       &       &  \\
    \cmidrule(l){4-5}
          &       &       & iBetaCOBL &       & Original &       & AGOBL &       & BetaCOBL &       & COOBL &       & EOBL  &       & GOBL  &  \\
    \cmidrule(l){4-5} \cmidrule(l){6-7} \cmidrule(l){8-9} \cmidrule(l){10-11} \cmidrule(l){12-13} \cmidrule(l){14-15} \cmidrule(l){16-17}
    d     & T0    & T1    & T2    & (T2-T1)/T0 & T2    & (T2-T1)/T0 & T2    & (T2-T1)/T0 & T2    & (T2-T1)/T0 & T2    & (T2-T1)/T0 & T2    & (T2-T1)/T0 & T2    & (T2-T1)/T0 \\
    \midrule
    50    & 65.0  & 1068.0  & 3873.2  & 43.2  & 3598.0  & 38.9  & 3575.8  & 38.6  & 7786.6  & 103.4  & 3804.8  & 42.1  & 3561.0  & 38.4  & 3541.2  & 38.0  \\
    \midrule
    \midrule
          &       &       &       &       &       &       & OBL   &       & OBLPGJ &       & OBLTVJR &       & QOBL  &       & QROBL &  \\
    \cmidrule(l){8-9} \cmidrule(l){10-11} \cmidrule(l){12-13} \cmidrule(l){14-15} \cmidrule(l){16-17}
          &       &       &       &       &       &       & T2    & (T2-T1)/T0 & T2    & (T2-T1)/T0 & T2    & (T2-T1)/T0 & T2    & (T2-T1)/T0 & T2    & (T2-T1)/T0 \\
    \midrule
          &       &       &       &       &       &       & 3564.6 & 38.4  & 3549.2 & 38.2  & 3550.8 & 38.2  & 3566.4 & 38.4  & 3656.4 & 39.8 \\
    \bottomrule
    \end{tabular}%
  \label{tab:EDEV_complexity_cec13_50}%
\end{table*}%

\begin{table*}[htbp]
  \tiny
  \centering
  \caption{Algorithm complexity for EDEV with OBL variants on CEC 2017 test suite at 50-$D$.}
    \begin{tabular}{ccccccccccccccccc}
    \toprule
          &       &       & EDEV &       &       &       &       &       &       &       &       &       &       &       &       &  \\
    \cmidrule(l){4-5}
          &       &       & iBetaCOBL &       & Original &       & AGOBL &       & BetaCOBL &       & COOBL &       & EOBL  &       & GOBL  &  \\
    \cmidrule(l){4-5} \cmidrule(l){6-7} \cmidrule(l){8-9} \cmidrule(l){10-11} \cmidrule(l){12-13} \cmidrule(l){14-15} \cmidrule(l){16-17}
    d     & T0    & T1    & T2    & (T2-T1)/T0 & T2    & (T2-T1)/T0 & T2    & (T2-T1)/T0 & T2    & (T2-T1)/T0 & T2    & (T2-T1)/T0 & T2    & (T2-T1)/T0 & T2    & (T2-T1)/T0 \\
    \midrule
    50    & 62.0  & 573.0  & 2571.4  & 32.2  & 2392.2  & 29.3  & 2387.4  & 29.3  & 6814.2  & 100.7  & 3889.2  & 53.5  & 2330.8  & 28.4  & 2357.6  & 28.8  \\
    \midrule
    \midrule
          &       &       &       &       &       &       & OBL   &       & OBLPGJ &       & OBLTVJR &       & QOBL  &       & QROBL &  \\
    \cmidrule(l){8-9} \cmidrule(l){10-11} \cmidrule(l){12-13} \cmidrule(l){14-15} \cmidrule(l){16-17}
          &       &       &       &       &       &       & T2    & (T2-T1)/T0 & T2    & (T2-T1)/T0 & T2    & (T2-T1)/T0 & T2    & (T2-T1)/T0 & T2    & (T2-T1)/T0 \\
    \midrule
          &       &       &       &       &       &       & 2421.2 & 29.8  & 2328.6 & 28.3  & 2343.6 & 28.6  & 2391.0 & 29.3  & 2638.4 & 33.3 \\ \\
    \bottomrule
    \end{tabular}%
  \label{tab:EDEV_complexity_cec17_50}%
\end{table*}%

\begin{figure*}[htbp]
 \centering
 \subfigure[$F_{10}$ in CEC 2013]{
  \includegraphics[scale=0.25]{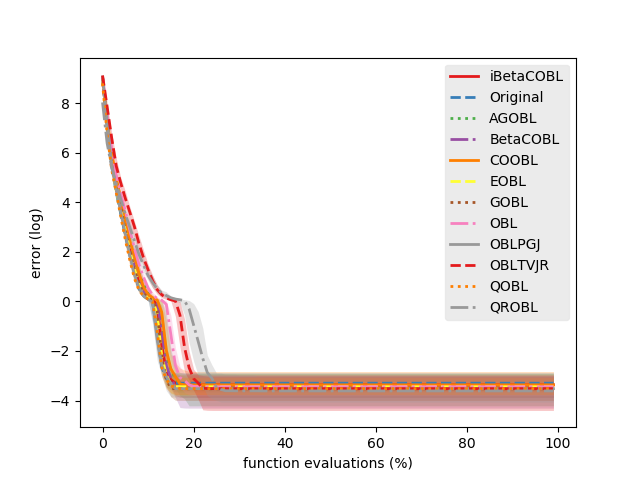}
   \label{fig:EDEV_f10_2013}
   }
 \subfigure[$F_{12}$ in CEC 2013]{
  \includegraphics[scale=0.25]{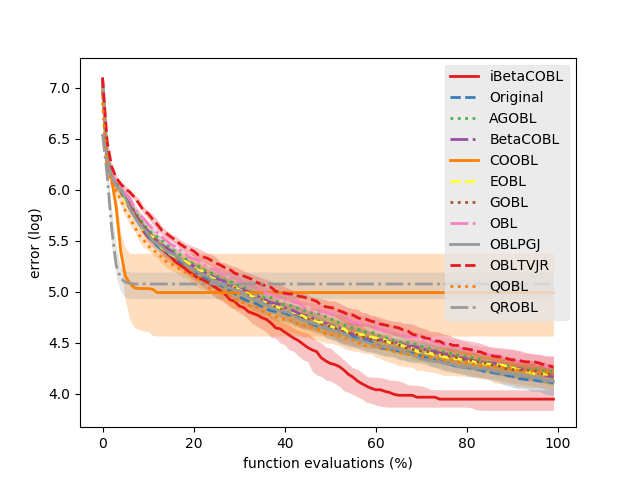}
   \label{fig:EDEV_f12_2013}
   }
 \subfigure[$F_{14}$ in CEC 2013]{
  \includegraphics[scale=0.25]{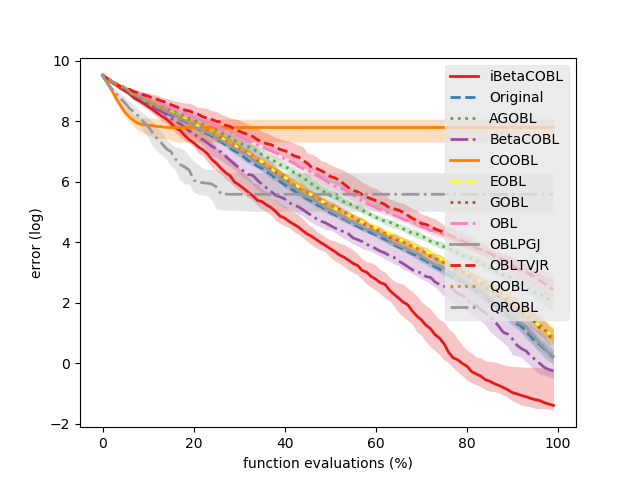}
   \label{fig:EDEV_f14_2013}
   }
 \subfigure[$F_{16}$ in CEC 2013]{
  \includegraphics[scale=0.25]{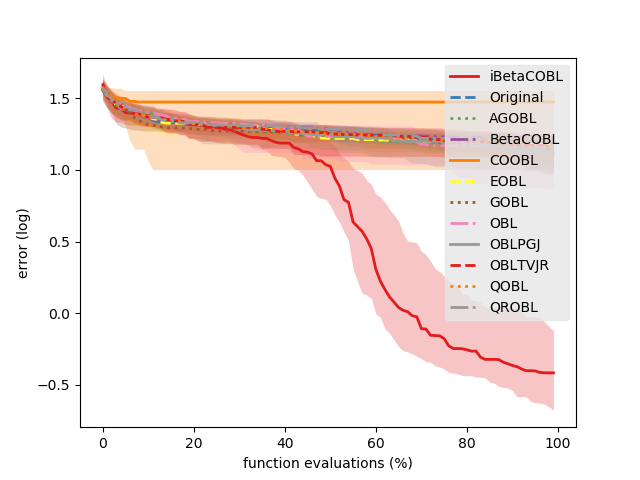}
   \label{fig:EDEV_f16_2013}
   }
 \subfigure[$F_{18}$ in CEC 2013]{
  \includegraphics[scale=0.25]{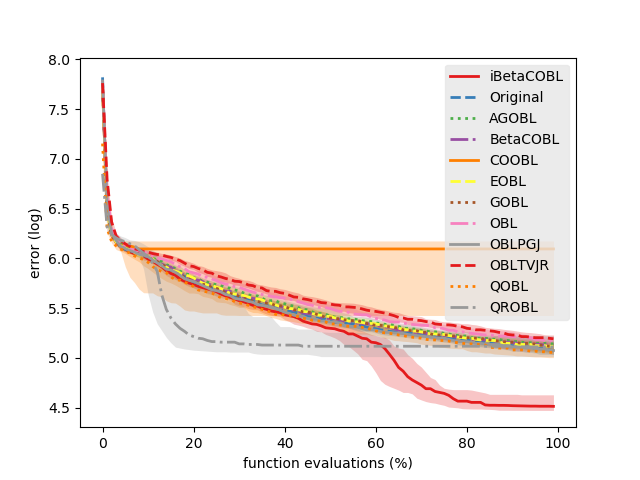}
   \label{fig:EDEV_f18_2013}
   }
 \subfigure[$F_{20}$ in CEC 2013]{
  \includegraphics[scale=0.25]{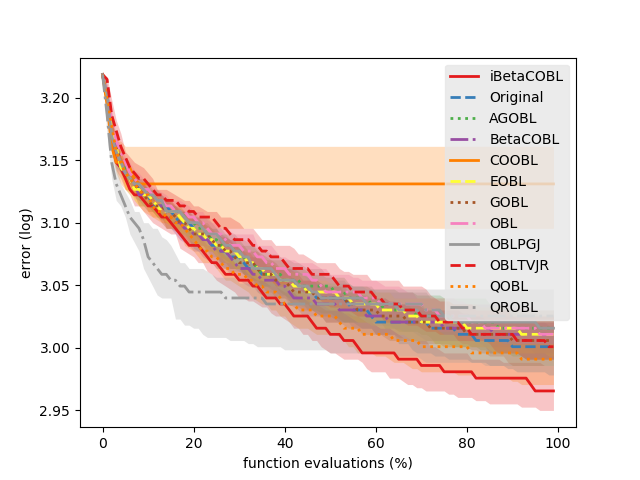}
   \label{fig:EDEV_f20_2013}
   }
 \subfigure[$F_{22}$ in CEC 2013]{
  \includegraphics[scale=0.25]{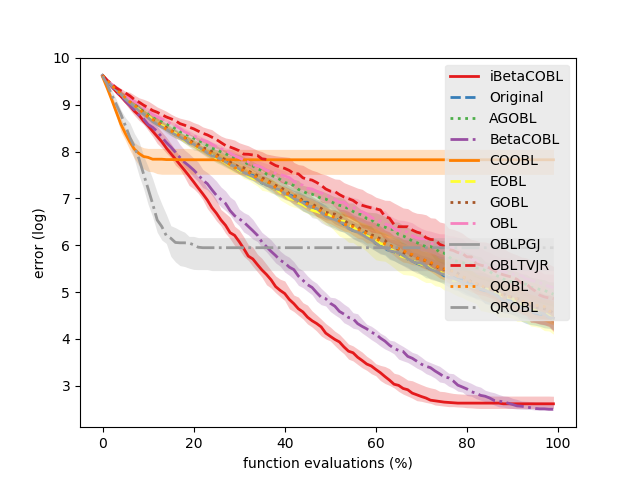}
   \label{fig:EDEV_f22_2013}
   }
 \subfigure[$F_{24}$ in CEC 2013]{
  \includegraphics[scale=0.25]{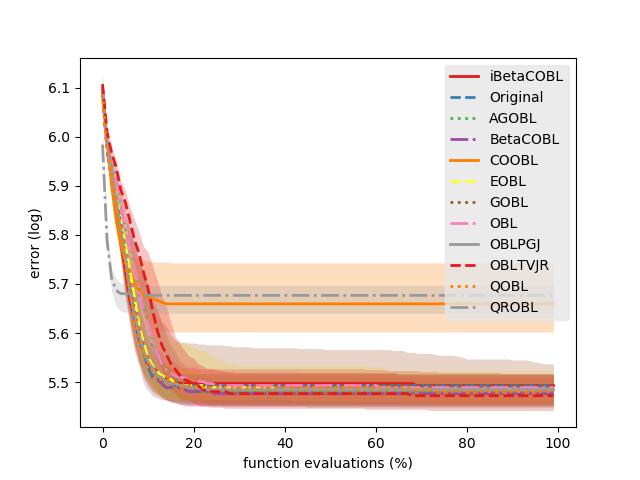}
   \label{fig:EDEV_f24_2013}
   }
 \subfigure[$F_{5}$ in CEC 2017]{
  \includegraphics[scale=0.25]{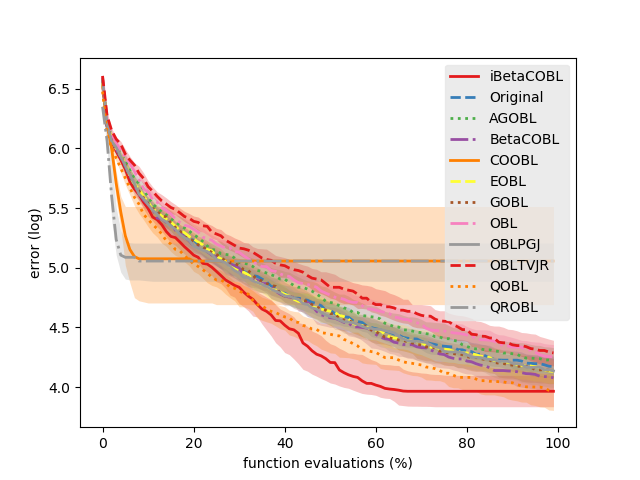}
   \label{fig:EDEV_f5_2017}
   }
 \subfigure[$F_{6}$ in CEC 2017]{
  \includegraphics[scale=0.25]{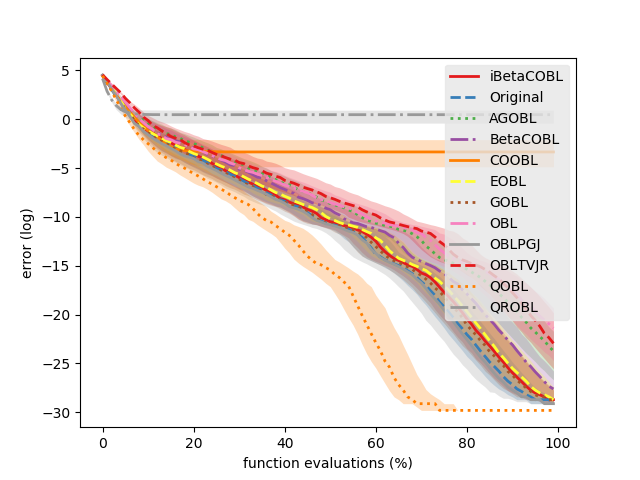}
   \label{fig:EDEV_f6_2017}
   }
 \subfigure[$F_{7}$ in CEC 2017]{
  \includegraphics[scale=0.25]{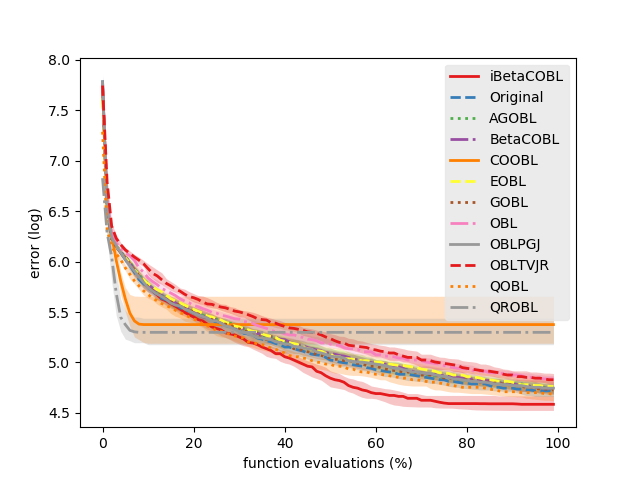}
   \label{fig:EDEV_f7_2017}
   }
 \subfigure[$F_{8}$ in CEC 2017]{
  \includegraphics[scale=0.25]{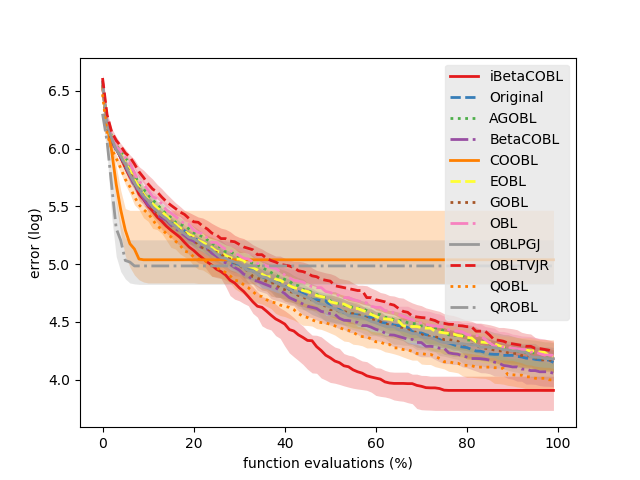}
   \label{fig:EDEV_f8_2017}
   }
 \subfigure[$F_{9}$ in CEC 2017]{
  \includegraphics[scale=0.25]{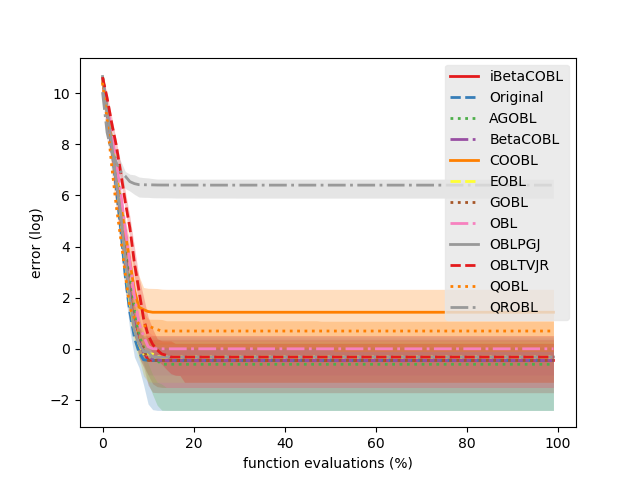}
   \label{fig:EDEV_f9_2017}
   }
 \subfigure[$F_{10}$ in CEC 2017]{
  \includegraphics[scale=0.25]{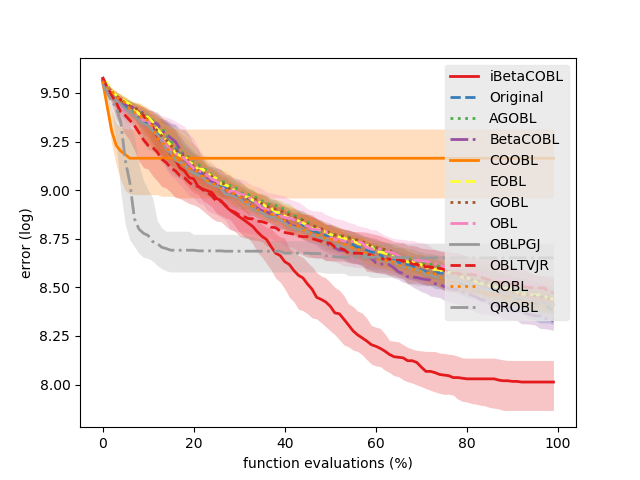}
   \label{fig:EDEV_f10_2017}
   }
 \subfigure[$F_{11}$ in CEC 2017]{
  \includegraphics[scale=0.25]{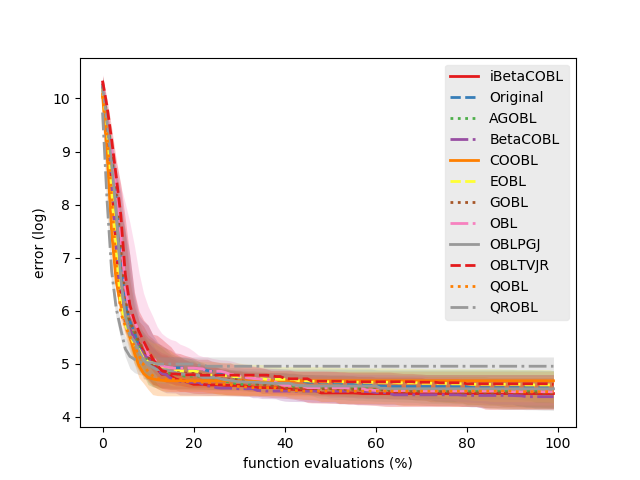}
   \label{fig:EDEV_f11_2017}
   }
 \subfigure[$F_{12}$ in CEC 2017]{
  \includegraphics[scale=0.25]{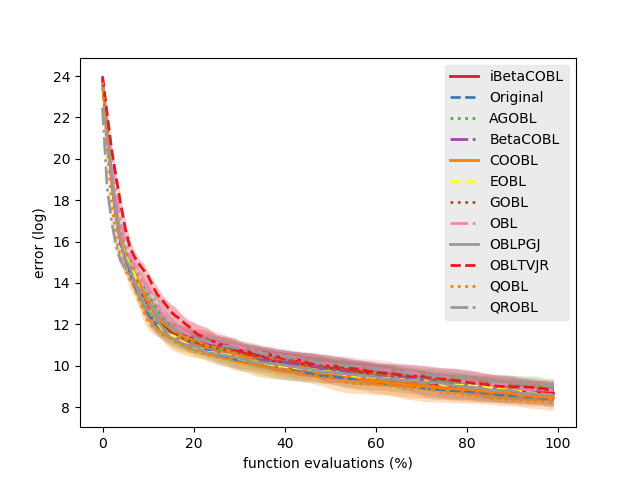}
   \label{fig:EDEV_f12_2017}
   }
 \caption[]{Convergence graphs of EDEV assisted by OBL variants on CEC 2013 and 2017 test suites at 50-$D$}
 \label{fig:EDEV_convergence}
\end{figure*}

The performance evaluation results of the EDEV variants on the CEC 2013 and 2017 test suites are presented in Tables \ref{tab:EDEV_wilcoxon_cec13} and \ref{tab:EDEV_wilcoxon_cec17}, as collected through 51 independent runs. As we can see from Table \ref{tab:EDEV_wilcoxon_cec13}, EDEV-iBetaCOBL points to promising overall performance compared to the other OBL variants on the CEC 2013 test suite. EDEV-iBetaCOBL found significantly better solutions than the other OBL variants on more than half of the benchmark problems. Also, the COOBL, OBL, OBLTVJR, and QROBL variants could not discover any better solution than the iBetaCOBL variant on all the benchmark problems. The results of the Friedman test with Hochberg's post hoc are presented in Table \ref{tab:EDEV_posthoc_cec13}, which supports the experimental results. According to the Friedman test, the original EDEV ranked the third among the test algorithms. Only EDEV assisted by iBetaCOBL and BetaCOBL ranked higher than the original EDEV, which is notable. We found similar tendencies on the CEC 2017 test suite in Tables \ref{tab:EDEV_wilcoxon_cec17} and \ref{tab:EDEV_posthoc_cec17}. In summary, the results of the performance evaluations show the excellent performance of the iBetaCOBL variant compared to the other OBL variants on both of the CEC 2013 and 2017 test suites.

Moreover, we analyzed the algorithm complexity of each algorithm. The results on the CEC 2013 and 2017 test suites are presented in Tables \ref{tab:EDEV_complexity_cec17_50} and \ref{tab:EDEV_complexity_cec13_50}, respectively. As we can see from the table, EDEV-iBetaCOBL consumed significantly less computational cost compared to EDEV-BetaCOBL. That is, the algorithm complexity of EDEV-BetaCOBL is approximately three times higher than EDEV-iBetaCOBL. On the other hand, the algorithm complexity of EDEV-iBetaCOBL is approximately similar or slightly higher than the other OBL variants.

Furthermore, Fig. \ref{fig:EDEV_convergence} presents the convergence graphs of the EDEV variants on 16 benchmark problems from the CEC 2013 and 2017 test suites. As we can see from the figures, the convergence progress of EDEV-iBetaCOBL is significantly better than that of the compared algorithms. Although the COOBL and QROBL variants have a faster convergence than the iBetaCOBL variant, it often fall into the local optimum. In particular, Figs. \ref{fig:EDEV_f12_2013}, \ref{fig:EDEV_f16_2013}, \ref{fig:EDEV_f18_2013}, \ref{fig:EDEV_f20_2013}, \ref{fig:EDEV_f5_2017}, \ref{fig:EDEV_f7_2017}, \ref{fig:EDEV_f8_2017}, \ref{fig:EDEV_f10_2017} show that EDEV-iBetaCOBL was able to escape the local optimum while the other OBL variants were not.

Consequently, we make the following observations on the performance evaluation results.

\begin{enumerate}
\item A significant performance improvement of EDEV can be achieved by incorporating the proposed OBL. \label{item:1}
\item EDEV-iBetaCOBL searched out more accurate solutions than EDEV-BetaCOBL with a significantly lower computational cost on the CEC 2013 and 2017 test suites. \label{item:2}
\item EDEV-iBetaCOBL shows promising convergence performance, with a better searchability than the other OBL variants. \label{item:3}
\end{enumerate}


\subsection{Performance Enhancement of LSHADE-RSP Algorithm}

\begin{table*}[htbp]
  \tiny
  \centering
  \caption{Averages and standard deviations of FEVs of LSHADE-RSP with OBL variants on CEC 2013 test suite at 50-$D$.}
\begin{tabular}{cccccccc}
    \toprule
          & LSHADE-RSP &       &       &       &       &       &  \\
    \cmidrule(l){2-2}
          & iBetaCOBL & Original & AGOBL & BetaCOBL & COOBL & EOBL  & GOBL \\
          & MEAN (STD DEV) & MEAN (STD DEV) & MEAN (STD DEV) & MEAN (STD DEV) & MEAN (STD DEV) & MEAN (STD DEV) & MEAN (STD DEV) \\
    \midrule
    F1    & 4.45E-15 (3.18E-14) & 0.00E+00 (0.00E+00) = & 0.00E+00 (0.00E+00) = & 0.00E+00 (0.00E+00) = & 0.00E+00 (0.00E+00) = & 4.45E-15 (3.18E-14) = & 0.00E+00 (0.00E+00) = \\
    F2    & 2.43E+00 (3.94E+00) & 7.50E+00 (4.31E+01) = & 7.16E+00 (2.38E+01) = & 2.34E+01 (7.80E+01) = & 7.34E+00 (2.11E+01) = & 5.92E+00 (1.49E+01) = & 9.88E+00 (3.25E+01) = \\
    F3    & 1.57E+00 (8.11E+00) & 1.05E+00 (4.53E+00) = & 5.50E+00 (2.40E+01) = & 3.07E+00 (1.65E+01) = & 1.04E+01 (6.61E+01) = & 5.31E+00 (1.60E+01) = & 6.03E+00 (2.64E+01) = \\
    F4    & 1.44E-08 (1.88E-08) & 2.07E-08 (2.54E-08) = & 6.99E-09 (1.07E-08) + & 1.10E-08 (1.19E-08) = & 6.67E-09 (6.47E-09) + & 7.28E-09 (6.46E-09) = & 1.01E-08 (1.95E-08) = \\
    F5    & 1.12E-13 (1.60E-14) & 1.14E-13 (7.65E-29) = & 1.12E-13 (1.60E-14) = & 1.14E-13 (7.65E-29) = & 1.12E-13 (1.60E-14) = & 2.24E+03 (1.14E+04) = & 1.14E-13 (7.65E-29) = \\
\bottomrule
    \end{tabular}%
  \label{tab:LSHADE-RSP_wilcoxon_cec13}%
\footnotesize
\\The symbols ``+/=/-" indicate that LSHADE-RSP with a given OBL performed significantly better ($+$), not significantly better or worse ($=$), or significantly worse ($-$) compared to LSHADE-RSP with iBetaCOBL using the Wilcoxon rank-sum test with $\alpha = 0.05$ significance level.
\end{table*}%

\begin{table*}[htbp]
  \tiny
  \centering
  \caption{Friedman test with Hochberg's post hoc for LSHADE-RSP with OBL variants on CEC 2013 test suite at 50-$D$.}
    \begin{tabular}{cccccccccc}
    \toprule
          & LSHADE-RSP &       &       &       &       &       &       &       &  \\
    \cmidrule(l){2-2}
          & Algorithm & Average ranking & z-value & p-value & Adj. p-value (Hochberg) & Sig.  &       & Test statistics &  \\
    \midrule
    1     & iBetaCOBL & 4.70  &       &       &       &       &       &       &  \\
    2     & Original & 5.25  & -5.74.E-01 & 5.66.E-01 & 5.66.E-01 & No    &       & N     & 28  \\
    3     & AGOBL & 7.79  & -3.21.E+00 & 1.35.E-03 & 1.35.E-02 & Yes   &       & Chi-Square & 26.35  \\
    4     & BetaCOBL & 6.86  & -2.24.E+00 & 2.49.E-02 & 1.75.E-01 & No    &       & df    & 11  \\
    5     & COOBL & 7.00  & -2.39.E+00 & 1.68.E-02 & 1.51.E-01 & No    &       & p-value & 5.75.E-03 \\
    6     & EOBL  & 6.96  & -2.35.E+00 & 1.86.E-02 & 1.49.E-01 & No    &       & Sig.  & Yes \\
    7     & GOBL  & 6.71  & -2.09.E+00 & 3.63.E-02 & 1.81.E-01 & No    &       &       &  \\
    8     & OBL   & 8.34  & -3.78.E+00 & 1.57.E-04 & 1.72.E-03 & Yes   &       &       &  \\
    9     & OBLPGJ & 6.77  & -2.15.E+00 & 3.16.E-02 & 1.90.E-01 & No    &       &       &  \\
    10    & OBLTVJR & 6.23  & -1.59.E+00 & 1.11.E-01 & 4.44.E-01 & No    &       &       &  \\
    11    & QOBL  & 5.30  & -6.30.E-01 & 5.29.E-01 & 1.06.E+00 & No    &       &       &  \\
    12    & QROBL & 6.09  & -1.45.E+00 & 1.48.E-01 & 4.45.E-01 & No    &       &       &  \\
    \bottomrule
    \end{tabular}%
  \label{tab:LSHADE-RSP_posthoc_cec13}%
\end{table*}%

\begin{table*}[htbp]
  \tiny
  \centering
  \caption{Averages and standard deviations of FEVs of LSHADE-RSP with OBL variants on CEC 2017 test suite at 50-$D$.}
    \begin{tabular}{cccccccc}
    \toprule
          & LSHADE-RSP &       &       &       &       &       &  \\
    \cmidrule(l){2-2}
          & iBetaCOBL & Original & AGOBL & BetaCOBL & COOBL & EOBL  & GOBL \\
          & MEAN (STD DEV) & MEAN (STD DEV) & MEAN (STD DEV) & MEAN (STD DEV) & MEAN (STD DEV) & MEAN (STD DEV) & MEAN (STD DEV) \\
    \midrule
    F1    & 1.45E-14 (3.47E-15) & 1.61E-14 (4.94E-15) = & 1.61E-14 (6.36E-15) = & 1.45E-14 (5.31E-15) = & 1.61E-14 (1.10E-14) = & 1.36E-14 (3.98E-15) = & 1.50E-14 (6.61E-15) = \\
    F2    & 7.86E-14 (1.29E-13) & 5.85E-14 (7.49E-14) = & 7.69E-14 (1.34E-13) = & 8.69E-14 (1.59E-13) = & 8.91E-14 (3.43E-13) = & 3.90E-14 (4.70E-14) = & 4.90E-14 (8.72E-14) = \\
    F3    & 1.39E-13 (4.44E-14) & 1.23E-13 (3.68E-14) = & 6.69E-14 (2.20E-14) + & 1.41E-13 (3.48E-14) = & 1.32E-13 (3.68E-14) = & 1.33E-13 (3.72E-14) = & 1.23E-13 (3.49E-14) = \\
    F4    & 5.94E+01 (4.71E+01) & 5.88E+01 (4.97E+01) = & 4.72E+01 (3.78E+01) = & 4.77E+01 (4.16E+01) = & 4.72E+01 (4.11E+01) = & 5.18E+01 (4.51E+01) = & 4.28E+01 (4.05E+01) + \\
    F5    & 1.49E+01 (3.47E+00) & 1.51E+01 (4.64E+00) = & 1.73E+01 (5.42E+00) - & 1.52E+01 (4.28E+00) = & 1.74E+01 (6.00E+00) - & 1.58E+01 (4.48E+00) = & 1.53E+01 (4.49E+00) = \\
\bottomrule
    \end{tabular}%
  \label{tab:LSHADE-RSP_wilcoxon_cec17}%
\footnotesize
\\The symbols ``+/=/-" indicate that LSHADE-RSP with a given OBL performed significantly better ($+$), not significantly better or worse ($=$), or significantly worse ($-$) compared to LSHADE-RSP with iBetaCOBL using the Wilcoxon rank-sum test with $\alpha = 0.05$ significance level.
\end{table*}%

\begin{table*}[htbp]
  \tiny
  \centering
  \caption{Friedman test with Hochberg's post hoc for LSHADE-RSP with OBL variants on CEC 2017 test suite at 50-$D$.}
    \begin{tabular}{cccccccccc}
    \toprule
          & LSHADE-RSP &       &       &       &       &       &       &       &  \\
    \cmidrule(l){2-2}
          & Algorithm & Average ranking & z-value & p-value & Adj. p-value (Hochberg) & Sig.  &       & Test statistics &  \\
    \midrule
    1     & iBetaCOBL & 5.47  &       &       &       &       &       &       &  \\
    2     & Original & 6.08  & -6.62.E-01 & 5.08.E-01 & 2.54.E+00 & No    &       & N     & 30  \\
    3     & AGOBL & 6.40  & -1.00.E+00 & 3.16.E-01 & 1.90.E+00 & No    &       & Chi-Square & 40.84  \\
    4     & BetaCOBL & 5.25  & 2.33.E-01 & 8.16.E-01 & 8.16.E-01 & No    &       & df    & 11  \\
    5     & COOBL & 8.88  & -3.67.E+00 & 2.42.E-04 & 2.67.E-03 & Yes   &       & p-value & 2.57.E-05 \\
    6     & EOBL  & 7.35  & -2.02.E+00 & 4.31.E-02 & 3.88.E-01 & No    &       & Sig.  & Yes \\
    7     & GOBL  & 5.70  & -2.51.E-01 & 8.02.E-01 & 1.60.E+00 & No    &       &       &  \\
    8     & OBL   & 8.57  & -3.33.E+00 & 8.69.E-04 & 8.69.E-03 & Yes   &       &       &  \\
    9     & OBLPGJ & 6.58  & -1.20.E+00 & 2.30.E-01 & 1.61.E+00 & No    &       &       &  \\
    10    & OBLTVJR & 5.73  & -2.86.E-01 & 7.75.E-01 & 2.32.E+00 & No    &       &       &  \\
    11    & QOBL  & 4.88  & 6.27.E-01 & 5.31.E-01 & 2.12.E+00 & No    &       &       &  \\
    12    & QROBL & 7.10  & -1.75.E+00 & 7.93.E-02 & 6.35.E-01 & No    &       &       &  \\
    \bottomrule
    \end{tabular}%
  \label{tab:LSHADE-RSP_posthoc_cec17}%
\end{table*}%

\begin{table*}[htbp]
  \tiny
  \centering
  \caption{Algorithm complexity for LSHADE-RSP with OBL variants on CEC 2013 test suite at 50-$D$.}
    \begin{tabular}{ccccccccccccccccc}
    \toprule
          &       &       & LSHADE-RSP &       &       &       &       &       &       &       &       &       &       &       &       &  \\
    \cmidrule(l){4-5}
          &       &       & iBetaCOBL &       & Original &       & AGOBL &       & BetaCOBL &       & COOBL &       & EOBL  &       & GOBL  &  \\
    \cmidrule(l){4-5} \cmidrule(l){6-7} \cmidrule(l){8-9} \cmidrule(l){10-11} \cmidrule(l){12-13} \cmidrule(l){14-15} \cmidrule(l){16-17}
    d     & T0    & T1    & T2    & (T2-T1)/T0 & T2    & (T2-T1)/T0 & T2    & (T2-T1)/T0 & T2    & (T2-T1)/T0 & T2    & (T2-T1)/T0 & T2    & (T2-T1)/T0 & T2    & (T2-T1)/T0 \\
    \midrule
    50    & 64.0  & 1081.0  & 18633.0  & 274.3  & 18684.8  & 275.1  & 18546.8  & 272.9  & 22178.4  & 329.6  & 18501.6  & 272.2  & 18590.2  & 273.6  & 18583.6  & 273.5  \\
    \midrule
    \midrule
          &       &       &       &       &       &       & OBL   &       & OBLPGJ &       & OBLTVJR &       & QOBL  &       & QROBL &  \\
    \cmidrule(l){8-9} \cmidrule(l){10-11} \cmidrule(l){12-13} \cmidrule(l){14-15} \cmidrule(l){16-17}
          &       &       &       &       &       &       & T2    & (T2-T1)/T0 & T2    & (T2-T1)/T0 & T2    & (T2-T1)/T0 & T2    & (T2-T1)/T0 & T2    & (T2-T1)/T0 \\
    \midrule
          &       &       &       &       &       &       & 18504.2 & 272.2 & 18605.8 & 273.8 & 18564.6 & 273.2 & 18615.2 & 274.0 & 18535.0 & 272.7 \\
    \bottomrule
    \end{tabular}%
  \label{tab:LSHADE-RSP_complexity_cec13_50}%
\end{table*}%

\begin{table*}[htbp]
  \tiny
  \centering
  \caption{Algorithm complexity for LSHADE-RSP with OBL variants on CEC 2017 test suite at 50-$D$.}
    \begin{tabular}{ccccccccccccccccc}
    \toprule
          &       &       & LSHADE-RSP &       &       &       &       &       &       &       &       &       &       &       &       &  \\
    \cmidrule(l){4-5}
          &       &       & iBetaCOBL &       & Original &       & AGOBL &       & BetaCOBL &       & COOBL &       & EOBL  &       & GOBL  &  \\
    \cmidrule(l){4-5} \cmidrule(l){6-7} \cmidrule(l){8-9} \cmidrule(l){10-11} \cmidrule(l){12-13} \cmidrule(l){14-15} \cmidrule(l){16-17}
    d     & T0    & T1    & T2    & (T2-T1)/T0 & T2    & (T2-T1)/T0 & T2    & (T2-T1)/T0 & T2    & (T2-T1)/T0 & T2    & (T2-T1)/T0 & T2    & (T2-T1)/T0 & T2    & (T2-T1)/T0 \\
    \midrule
    50    & 61.0  & 581.0  & 17149.6  & 271.6  & 17236.2  & 273.0  & 17102.6  & 270.8  & 20821.2  & 331.8  & 17072.4  & 270.4  & 17143.4  & 271.5  & 17149.4  & 271.6  \\
    \midrule
    \midrule
          &       &       &       &       &       &       & OBL   &       & OBLPGJ &       & OBLTVJR &       & QOBL  &       & QROBL &  \\
    \cmidrule(l){8-9} \cmidrule(l){10-11} \cmidrule(l){12-13} \cmidrule(l){14-15} \cmidrule(l){16-17}
          &       &       &       &       &       &       & T2    & (T2-T1)/T0 & T2    & (T2-T1)/T0 & T2    & (T2-T1)/T0 & T2    & (T2-T1)/T0 & T2    & (T2-T1)/T0 \\
    \midrule
          &       &       &       &       &       &       & 17042.8 & 269.9 & 17170.2 & 272.0 & 17114.0 & 271.0 & 17135.4 & 271.4 & 17091.8 & 270.7 \\
    \bottomrule
    \end{tabular}%
  \label{tab:LSHADE-RSP_complexity_cec17_50}%
\end{table*}%

\begin{figure*}[htbp]
 \centering
 \subfigure[$F_{10}$ in CEC 2013]{
  \includegraphics[scale=0.25]{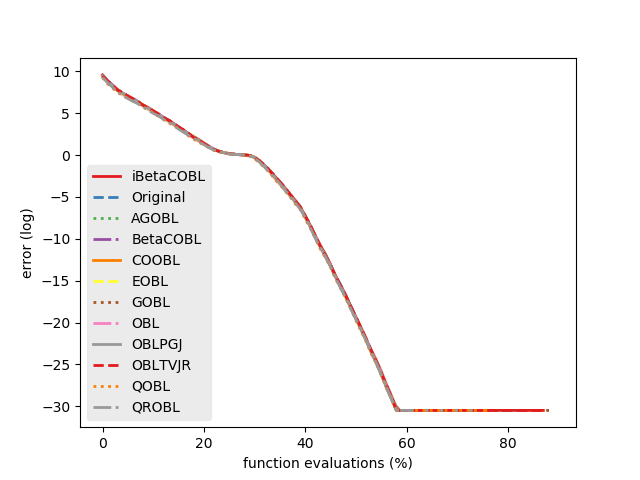}
   \label{fig:LSHADE-RSP_f10_2013}
   }
 \subfigure[$F_{12}$ in CEC 2013]{
  \includegraphics[scale=0.25]{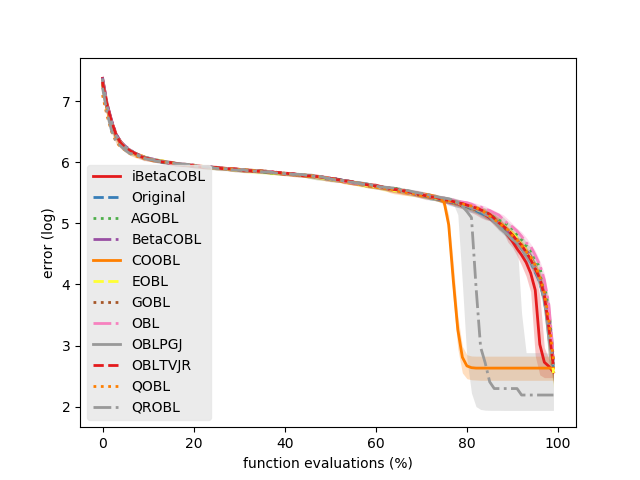}
   \label{fig:LSHADE-RSP_f12_2013}
   }
 \subfigure[$F_{14}$ in CEC 2013]{
  \includegraphics[scale=0.25]{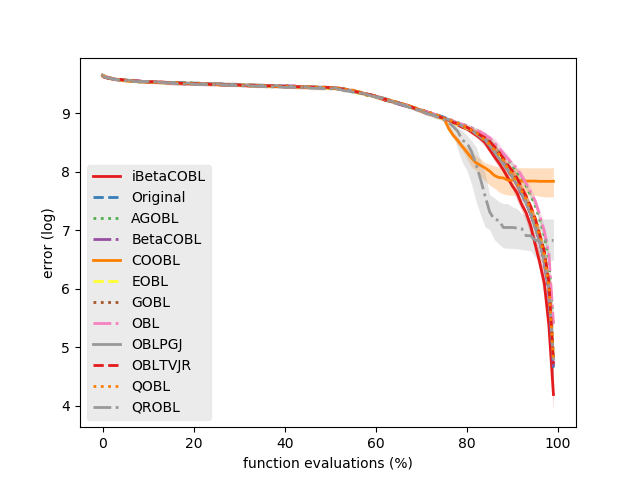}
   \label{fig:LSHADE-RSP_f14_2013}
   }
 \subfigure[$F_{16}$ in CEC 2013]{
  \includegraphics[scale=0.25]{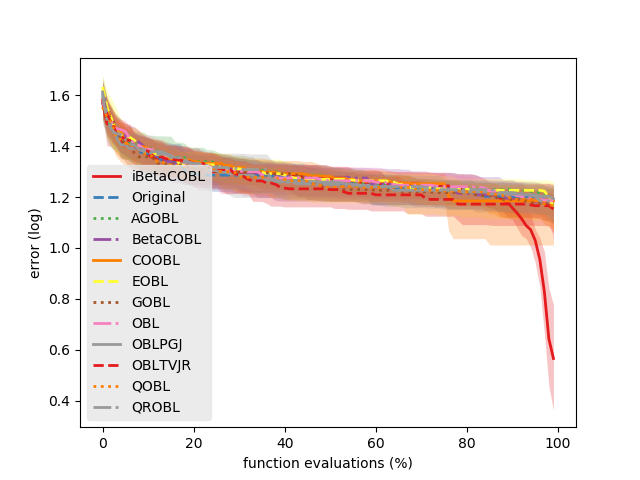}
   \label{fig:LSHADE-RSP_f16_2013}
   }
 \subfigure[$F_{18}$ in CEC 2013]{
  \includegraphics[scale=0.25]{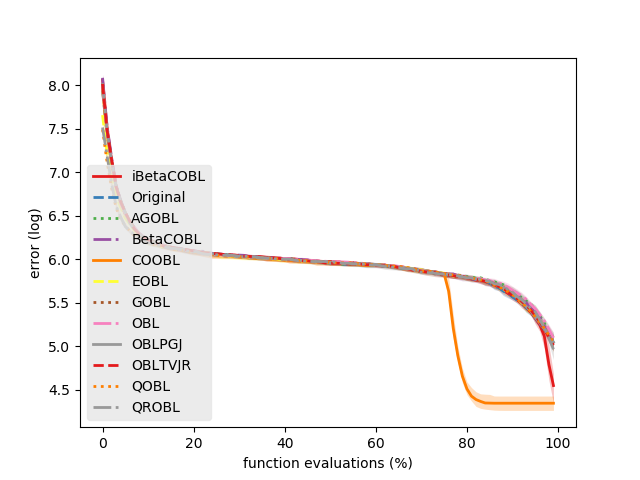}
   \label{fig:LSHADE-RSP_f18_2013}
   }
 \subfigure[$F_{20}$ in CEC 2013]{
  \includegraphics[scale=0.25]{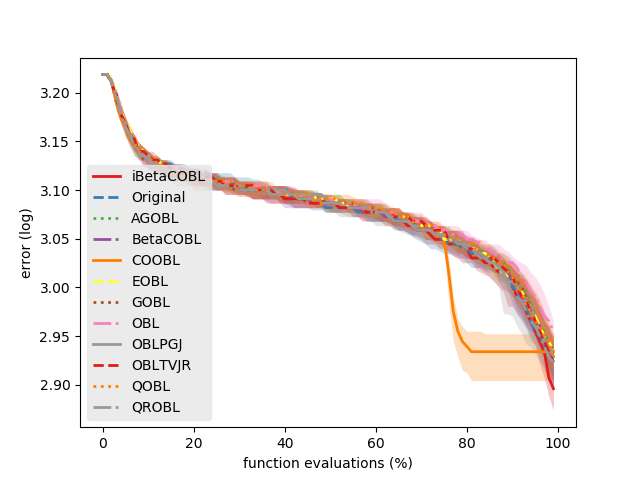}
   \label{fig:LSHADE-RSP_f20_2013}
   }
 \subfigure[$F_{22}$ in CEC 2013]{
  \includegraphics[scale=0.25]{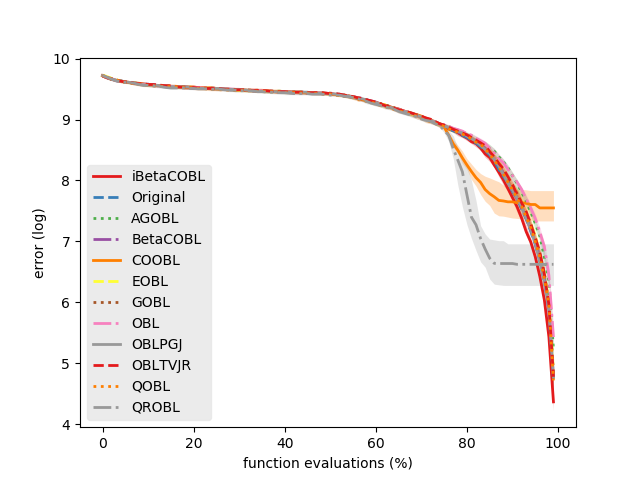}
   \label{fig:LSHADE-RSP_f22_2013}
   }
 \subfigure[$F_{24}$ in CEC 2013]{
  \includegraphics[scale=0.25]{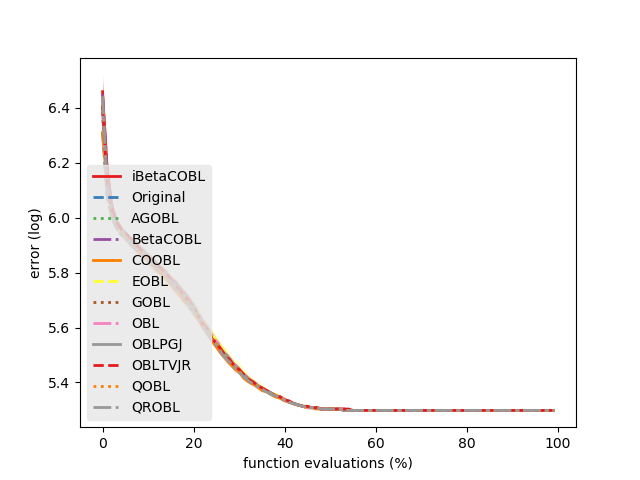}
   \label{fig:LSHADE-RSP_f24_2013}
   }
 \subfigure[$F_{5}$ in CEC 2017]{
  \includegraphics[scale=0.25]{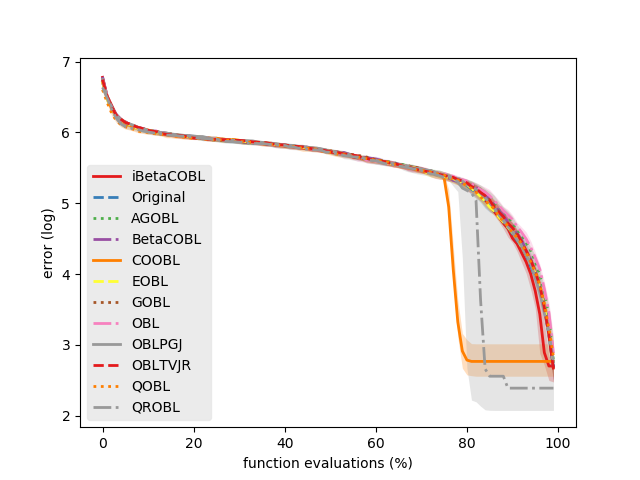}
   \label{fig:LSHADE-RSP_f5_2017}
   }
 \subfigure[$F_{6}$ in CEC 2017]{
  \includegraphics[scale=0.25]{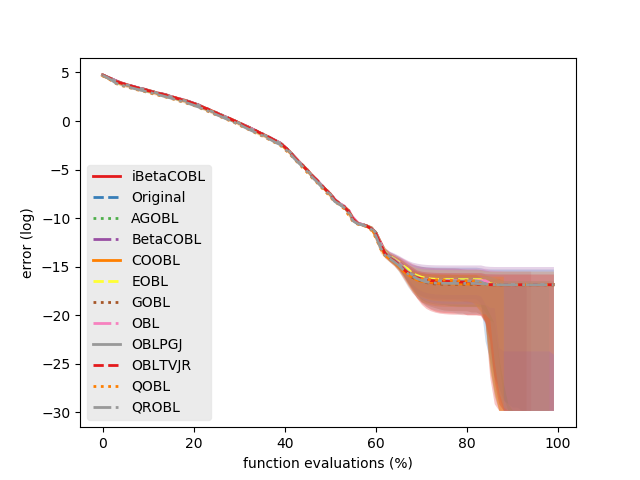}
   \label{fig:LSHADE-RSP_f6_2017}
   }
 \subfigure[$F_{7}$ in CEC 2017]{
  \includegraphics[scale=0.25]{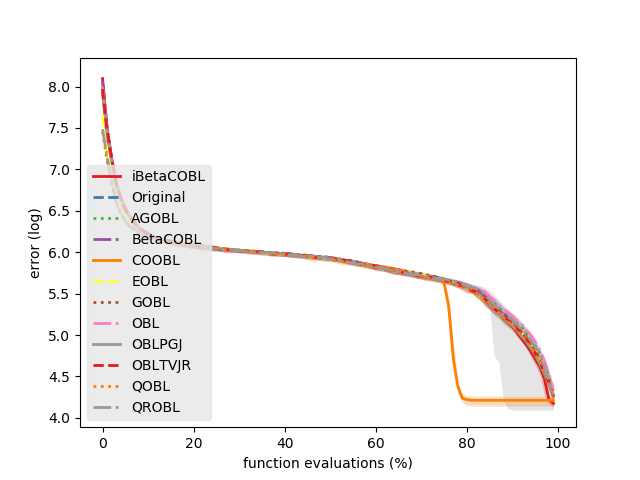}
   \label{fig:LSHADE-RSP_f7_2017}
   }
 \subfigure[$F_{8}$ in CEC 2017]{
  \includegraphics[scale=0.25]{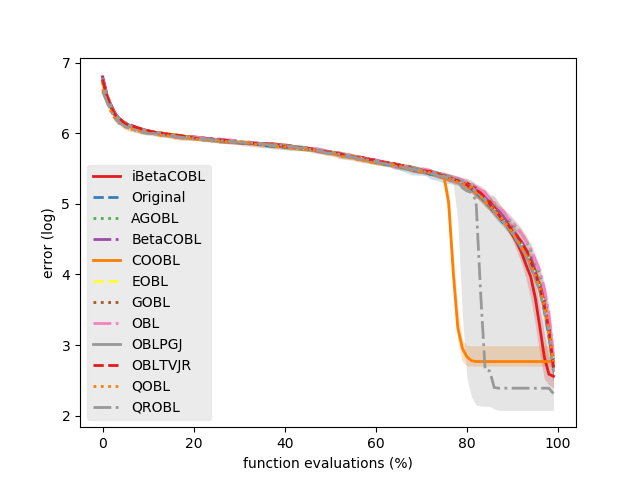}
   \label{fig:LSHADE-RSP_f8_2017}
   }
 \subfigure[$F_{9}$ in CEC 2017]{
  \includegraphics[scale=0.25]{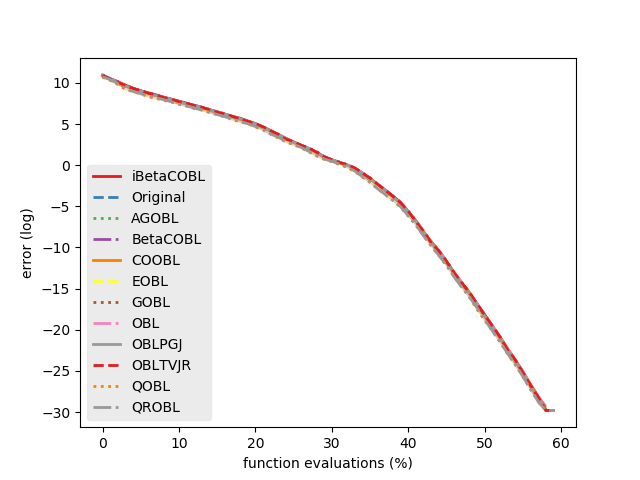}
   \label{fig:LSHADE-RSP_f9_2017}
   }
 \subfigure[$F_{10}$ in CEC 2017]{
  \includegraphics[scale=0.25]{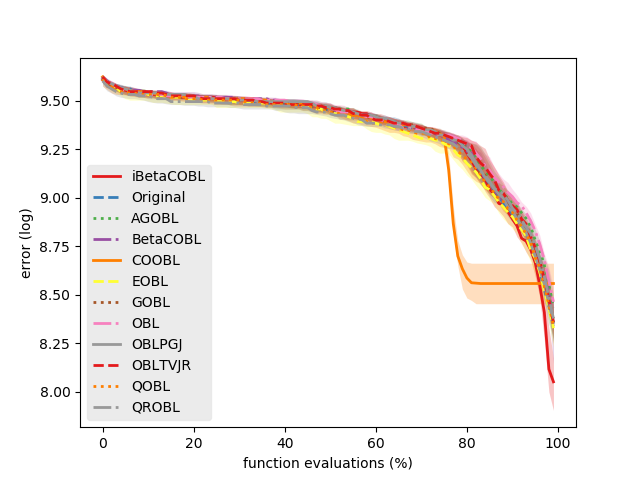}
   \label{fig:LSHADE-RSP_f10_2017}
   }
 \subfigure[$F_{11}$ in CEC 2017]{
  \includegraphics[scale=0.25]{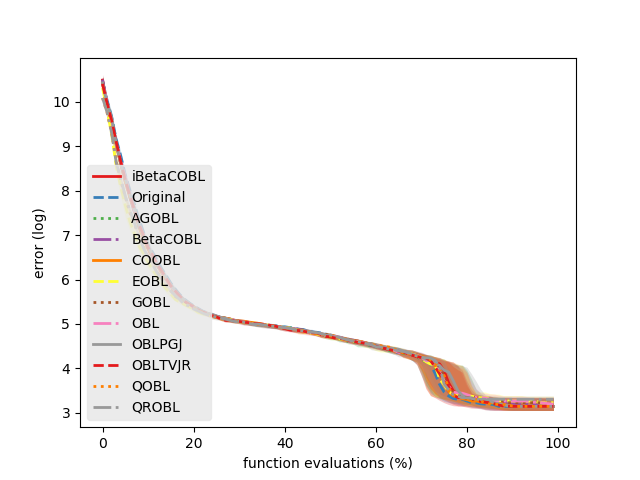}
   \label{fig:LSHADE-RSP_f11_2017}
   }
 \subfigure[$F_{12}$ in CEC 2017]{
  \includegraphics[scale=0.25]{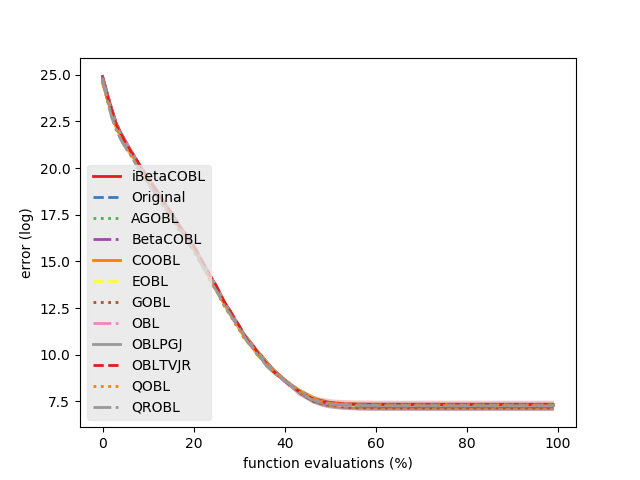}
   \label{fig:LSHADE-RSP_f12_2017}
   }
 \caption[]{Convergence graphs of LSHADE-RSP assisted by OBL variants on CEC 2013 and 2017 test suites at 50-$D$}
 \label{fig:LSHADE-RSP_convergence}
\end{figure*}

The results of performance evaluations of the LSHADE-RSP variants on the CEC 2013 and 2017 test suites are showed in Tables \ref{tab:LSHADE-RSP_wilcoxon_cec13} and \ref{tab:LSHADE-RSP_wilcoxon_cec17}, as obtained from 51 independent runs. LSHADE-RSP-iBetaCOBL points to impressive overall performance as opposed to the other OBL variants on the CEC 2013 test suite, as we can see from the table. In particular, for the multimodal and composition functions, LSHADE-RSP-iBetaCOBL found significantly better solutions than the other OBL variants. The BetaCOBL variant was also unable to discover any better solution on all the benchmark problems than the iBetaCOBL variant. The results of the Friedman test with Hochberg's post hoc are presented in Table \ref{tab:LSHADE-RSP_posthoc_cec13}, which supports the experimental results. According to the Friedman test, the original LSHADE-RSP ranked the second among the test algorithms. Only LSHADE-RSP assisted by iBetaCOBL ranked higher than the original LSHADE-RSP, which is notable. We found similar tendencies on the CEC 2017 test suite in Tables \ref{tab:LSHADE-RSP_wilcoxon_cec17} and \ref{tab:LSHADE-RSP_posthoc_cec17}. It should be noted that the average ranking of the BetaCOBL and QOBL variants is higher than the iBetaCOBL variant on the CEC 2017 test suite. However, the iBetaCOBL variant outperformed the BetaCOBL variant on the four test functions. Similarly, the iBetaCOBL variant outperformed the QOBL variant on the four test functions. the results of the performance evaluations show the excellent performance of the iBetaCOBL variant compared to the other OBL variants on both of the CEC 2013 and 2017 test suites.

Moreover, we analyzed the algorithm complexity of each algorithm. The results on the CEC 2013 and 2017 test suites are presented in Tables \ref{tab:LSHADE-RSP_complexity_cec17_50} and \ref{tab:LSHADE-RSP_complexity_cec13_50}, respectively. As we can see from the table, LSHADE-RSP-iBetaCOBL consumed significantly less computational cost compared to LSHADE-RSP-BetaCOBL. That is, the algorithm complexity of LSHADE-RSP-BetaCOBL is approximately 20 percent higher than LSHADE-RSP-iBetaCOBL. On the other hand, the algorithm complexity of LSHADE-RSP-iBetaCOBL is approximately similar or slightly higher than the other OBL variants.

Furthermore, Fig. \ref{fig:LSHADE-RSP_convergence} presents the convergence graphs of the LSHADE-RSP variants on 16 benchmark problems from the CEC 2013 and 2017 test suites. It should be noted that LSAHDE-RSP starts to run OBL when it reaches three-fourths of the maximum number of function evaluations. Therefore, the convergence graphs of the LSHADE-RSP variants are the same until they start to run OBL. As we can see from the figures, the convergence progress of LSHADE-RSP-iBetaCOBL is significantly better than that of the compared algorithms. Although the COOBL and QROBL variants have a faster convergence than the iBetaCOBL variant, they often fall into the local optimum. In particular, Figs. \ref{fig:LSHADE-RSP_f16_2013} and \ref{fig:LSHADE-RSP_f10_2017} show that LSHADE-RSP-iBetaCOBL was able to escape the local optimum while the other OBL variants were not.

Consequently, we make the following observations on the performance evaluation results.

\begin{enumerate}
\item A significant performance improvement of LSHADE-RSP can be achieved by incorporating the proposed OBL. \label{item:1}
\item LSHADE-RSP-iBetaCOBL searched out more accurate solutions than LSHADE-RSP-BetaCOBL with a significantly lower computational cost on the CEC 2013 and 2017 test suites. \label{item:2}
\item LSHADE-RSP-iBetaCOBL shows promising convergence performance, with a better searchability than the other OBL variants. \label{item:3}
\end{enumerate}


\section{Conclusion}
\label{sec:Conclusion}

We have proposed a cutting-edge OBL variant called iBetaCOBL, which is an improved BetaCOBL. Although it is a powerful OBL variant to accelerate the convergence of EAs, the main limitations of BetaCOBL are 1) high computational cost and 2) ineffectiveness in handling dependent decision variables. Because of the limitations, BetaCOBL to optimize cost-sensitive optimization problems or more complex problems may be impractical. The goal of this paper is to propose an advance OBL variant that mitigates all the limitations. To reduce the computational cost, we applied a linear time diversity measure in the selection switching scheme. Also, we applied multiple exponential crossover in the partial dimensional change scheme to preserve structures with strongly dependent decision variables adjacent to each other.

The performance of iBetaCOBL was evaluated on a set of 58 different and difficult test functions from the CEC 2013 and 2017 test suites. Our experiments confirm that iBetaCOBL has the ability to find more accurate solutions than ten state-of-the-art OBL variants. The most remarkable result to emerge from the experiments is that iBetaCOBL significantly outperformed its predecessor BetaCOBL with considerably less time complexity. Therefore, iBetaCOBL is a clear improvement on BetaCOBL. We also applied iBetaCOBL to two state-of-the-art DE variants, EDEV and LSAHDE-RSP, to investigate the compatibility. Consequently, we confirm that a significant performance improvement for the DE variants can be achieved using the proposed algorithm.

Possible directions for future work include 1) devising a Cauchy or Gaussian distribution-based OBL; 2) applying iBetaCOBL to multi-objective EAs; 3) analyzing the proposed algorithm using dynamic systems to prove and explain the convergence of the proposed algorithm.




\section*{Acknowledgement}

This work was supported by the National Research Foundation of Korea(NRF) grant funded by the Korea government(MSIT) (No. NRF-2017R1C1B2012752). The correspondence should be addressed to Dr. Yun-Gyung Cheong.

\bibliographystyle{elsarticle-num} 
\bibliography{refs}





\end{document}